%% file: TCAD_main.tex
\documentclass[journal]{IEEEtran}
\input{macro}
\usepackage{xcolor}
\usepackage{soul}

\ifCLASSINFOpdf
\else
\fi
\begin{document}
	\title{Exploring ``Many in Few'' and ``Few in Many'' Properties in Long-Tailed, Highly-Imbalanced IC Defect Classification}
	
	\author{Hao-Chiang~Shao,~\IEEEmembership{Senior Member,~IEEE}, Chun-Hao~Chang, Yu-Hsien~Lin, Chia-Wen~Lin,~\IEEEmembership{Fellow,~IEEE}, Shao-Yun~Fang,~\IEEEmembership{Senior Member,~IEEE}, 
	and~Yan-Hsiu~Liu% <-this % stops a space
	\thanks{Manuscript received on March 13, 2025. This work was supported in part by the National Council of Science and Technology, Taiwan, under Grants NSTC 113-2221-E-005-070-MY3, 113-2634-F-002-003, and 112-2221-E-007-077-MY3, and in part by United Microelectronics Corporation. (Corresponding Author: Chia-Wen Lin)}
	\thanks{Hao-Chiang Shao is with the Institute of Data Science
            and Information Computing, National Chung Hsing University, Taichung 402202, Taiwan. (e-mail:shao.haochiang@gmail.com)}
%	\thanks{Chun-Hao Chang is with the Department of Electrical Engineering, National Tsing Hua University, Hsinchu 300044, Taiwan.}		
	\thanks{Chun-Hao Chang, Yu-Hsien Lin, and Chia-Wen Lin are with the Department of Electrical Engineering, National Tsing Hua University, Hsinchu 300044, Taiwan. (e-mail: cwlin@ee.nthu.edu.tw)}
	\thanks{Shao-Yun Fang is with the Department of Electrical Engineering, National Taiwan University of Science and Technology, Taipei 106335, Taiwan. (e-mail: syfang@mail.ntust.edu.tw)}
%	\thanks{Pin-Yian Tsai is with the Product Engineering Department, United Microelectronics Corporation, Hsinchu 300094, Taiwan. (e-mail: pin\_yian\_tsai@umc.com)}
        \thanks{Yan-Hsiu Liu is with the Development of Smart Manufacturing, United Microelectronics Corporation, Hsinchu 300094, Taiwan. (e-mail: cecil\_liu@umc.com)}
	\thanks{Color versions of one or more of the figures in this paper are available online at http://ieeexplore.ieee.org.}
	}
	

	\markboth{IEEE Transactions on Computer-Aided Design of Integrated Circuits and Systems,~Vol.~x, No.~x, Month~2025}%
	{Shell \MakeLowercase{\textit{et al.}}: Bare Demo of IEEEtran.cls for IEEE Journals}



	\maketitle
	
	\begin{abstract}
        \input{TCAD_sec00_abstract}

	\end{abstract}
	

	% \fi
	\IEEEpeerreviewmaketitle

%------------------------------------------------------------------------	
	\section{Introduction}
	\label{sec01:intro}

\input{TCAD_sec01_introduction}
	\section{Related Work}
	\label{sec02:review}

\input{TCAD_sec02_review}

%------------------------------------------------------------------------	
	\section{Proposed Method}
	\label{sec03:method}

\input{TCAD_sec03_method}

%------------------------------------------------------------------------	
	\section{Experimental Results}
	\label{sec04:exp}
	\input{TCAD_sec04_experiment}
	
%------------------------------------------------------------------------	
	\section{Conclusion}
	\label{sec:conclusion}
	\input{TCAD_sec05_conclusion}

	
	
	\bibliographystyle{IEEEtran}
	\bibliography{TCAD_sec99_reference}
	
	\input{bio}

%	\input{bio}
	
	
%以下幾行先註解掉,
%要產生listoffigure和listoftable時才會需要用到
\iffalse	
\onecolumn

\listoffigures
\newpage

\listoftables
\newpage
\fi

\end{document}

%% file: macro.tex
\usepackage{color}
\usepackage{bbding}
\usepackage{pifont}
\usepackage{amssymb}
\usepackage{amsthm}
\usepackage{booktabs}
\usepackage{multirow} 
\usepackage{setspace}
\usepackage{epsfig}
\usepackage{graphicx}
\usepackage{subfigure}
\usepackage{algorithm}
\usepackage{algpseudocode}
\usepackage{url}
\usepackage{amsmath}
\usepackage{bm}
\usepackage{tablefootnote}
\usepackage{float}
\usepackage{colortbl} % 提供顏色設定

\usepackage{array}
\newcolumntype{L}[1]{>{\raggedright\let\newline\\\arraybackslash\hspace{0pt}}m{#1}}
\newcolumntype{C}[1]{>{\centering\let\newline\\\arraybackslash\hspace{0pt}}m{#1}}
\newcolumntype{R}[1]{>{\raggedleft\let\newline\\\arraybackslash\hspace{0pt}}m{#1}}

%% file: TCAD_sec00_abstract.tex
Despite significant advancements in deep classification techniques and in-lab automatic optical inspection (AOI) models for long-tailed or highly imbalanced data, applying these approaches to real-world IC defect classification tasks remains challenging. This difficulty stems from two primary factors. 
First, real-world conditions, such as the high yield-rate requirements in the IC industry, result in data distributions that are far more skewed than those found in general public imbalanced datasets. Consequently, classifiers designed for open imbalanced datasets often fail to perform effectively in real-world scenarios. Second, real-world samples exhibit a mix of class-specific attributes (e.g., defect types) and class-agnostic, domain-related features (e.g., design characteristics of product lines). This complexity adds significant difficulty to the classification process, particularly for highly imbalanced datasets. To address these challenges, this paper introduces the \textbf{IC-Defect-14} dataset, a large, highly imbalanced IC defect image dataset sourced from AOI systems deployed in real-world IC production lines. This dataset is characterized by its unique ``intra-class clusters'' property, which presents two major challenges: large intra-class diversity and high inter-class similarity. These characteristics, rarely found simultaneously in existing public datasets, significantly degrade the performance of current state-of-the-art classifiers for highly imbalanced data. 
To tackle this challenge, we propose the \textbf{Regional Channel Attention-based Multi-Expert Network} (\textbf{ReCAME-Net}). This network follows a multi-expert classifier framework and integrates a regional channel attention module, metric learning losses, a hard category mining strategy, and a knowledge distillation procedure. Extensive experimental evaluations demonstrate that ReCAME-Net outperforms previous state-of-the-art models on the \textbf{IC-Defect-14} dataset while maintaining comparable performance and competitiveness on general public datasets. Our resources can be found at \textcolor{gray}{https://github.com/YoursEver/ReCAME--Net}.

%% file: TCAD_sec01_introduction.tex
\begin{figure}
    \centering
    \includegraphics[width=0.48\textwidth]{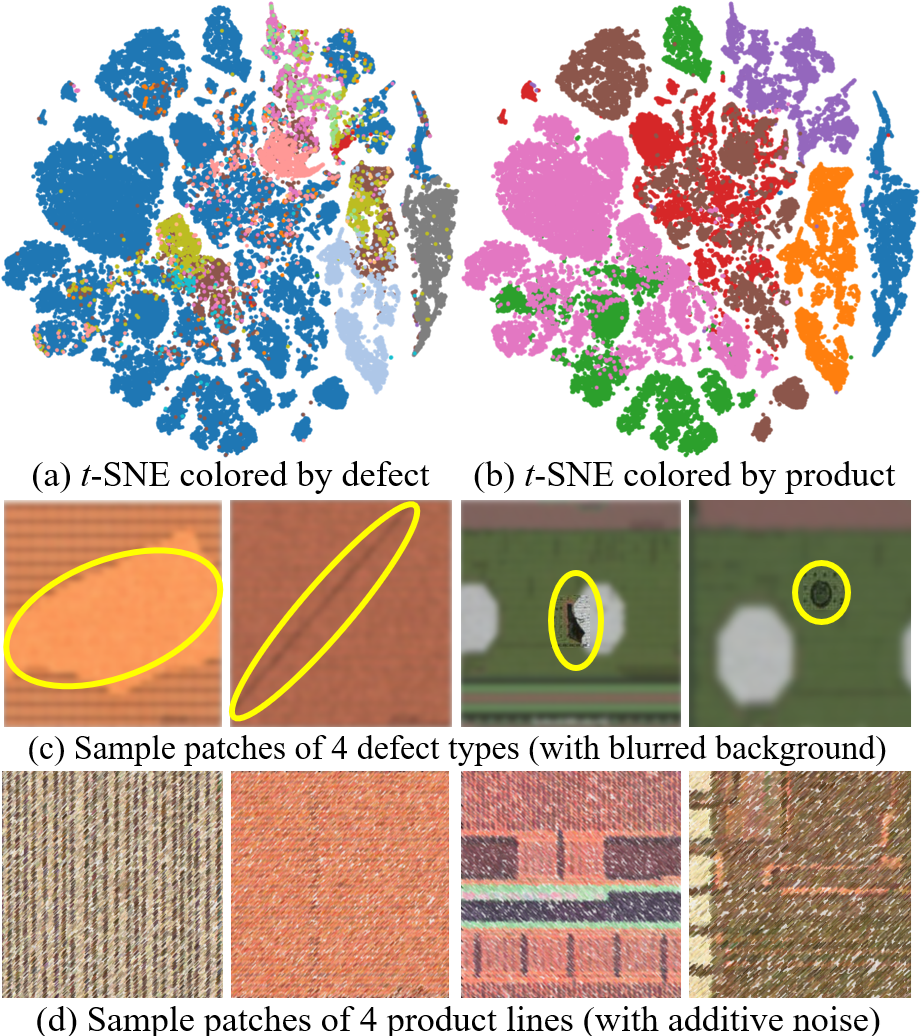}
    \caption{Challenges posed by a real-world highly-imbalanced IC defect image dataset. IC defect features are mixed with IC product design features, resulting in unexplored unique data properties that complicate the defect classification task. (a) The ``many-in-few''  property (i.e., \textit{multi-cluster head classes}) stems from significant \textit{intra-class diversity} in the feature space, where a single defect category exhibits multiple distinct clusters.   (b) Conversely, tail classes often contain very few samples, leading to the ``few-in-many'' property, where a sparse number of samples are scattered across a large feature space. Figures (c) and (d) demonstrate samples of different defect categories and product lines, highlighting that these data properties arise from different defect patterns on the same IC design and the same defect pattern on different IC designs. These unique data properties, combining diverse defect patterns and overlapping design influences, significantly complicate the classification process.}
   % }
    \label{fig:teaser}
\end{figure}

\begin{figure*}[!t]
    \centering
    \includegraphics[width=0.90\textwidth]{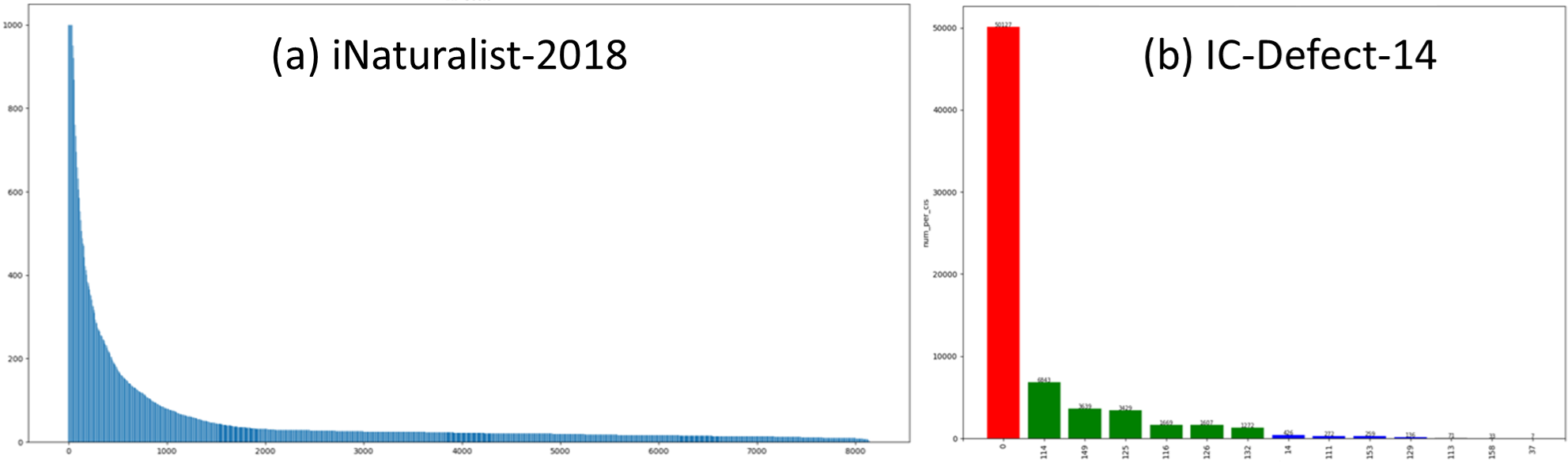}
    %{my_image/dis.png}
    %\includegraphics[width=\linewidth]{Proposed DE-GAN.png}
    %\captionsetup{font={bf,large}}
    %\captionsetup{font={large}}
    \caption{Data distributions of two imbalanced datasets: (a) iNaturalist-2018 \cite{van2018inaturalist} following an ideal exponential distribution, (b) ID-Defect-14 being extremely imbalanced. The vertical-axis and horizontal-axis are number of per-class samples and class indices, respectively.}
    \label{fig02:distribution_comparison}
\end{figure*}

\IEEEPARstart{A}{utomatic} defect classification is a widely adopted computer vision technique in the semiconductor industry, particularly for IC wafer post-manufacture quality assessment~\cite{abd2020review}. Despite significant advancements in classification techniques and in-lab automatic optical inspection (AOI) models for IC defect images, deploying these methods in real-world IC defect classification scenarios remains challenging due to two major constraints. First, because of the stringent yield-rate requirements in the IC industry, there is typically a vast number of normal samples and very few defect samples, leading to a highly imbalanced data distribution that complicates the learning of a deep classification model~\cite{liu2023machine}.  Second, real-world IC image samples exhibit a mix of class-specific attributes (e.g., defect types) and class-agnostic, domain-related features (e.g., design characteristics of product lines). For example, IC images of the same defect category (e.g., short circuits in the metal layer) may exhibit distinct outlook appearances due to the design differences of different product lines. The intertwining of i) product-related but defect-agnostic characteristics derived from IC design appearances, and ii) defect-specific appearances, further complicates the defect classification process, especially for highly class-imbalanced data distributions. 
%create an extremely imbalanced data distribution, characterized by an overwhelming number of normal samples compared to a limited number of defect samples. This imbalance hinders the effective training of deep classification models~\cite{liu2023machine}.  Second, defect images from different production lines can fall into the same defect category, while images within a single defect category may originate from diverse production lines.  This intertwining between product-specific characteristics and defect appearances introduces additional complexity, further exacerbating the challenges posed by the already imbalanced data distribution. 

%For example, it is challenging for existing long-tailed defect classification methods to handle the IC image defect dataset \textbf{IC-Defect-14} we investigate in this paper. As shown in Table~\ref{tab:results_on_UMC} in Section \ref{sec04:exp}, conventional ResNet-50~\cite{he2016deep} works compatibly with recent SOTAs of highly-imbalanced tasks on the \textbf{IC-Defect-14} dataset, implying recent SOTAs' insufficiency in handling the real-world data characteristics held by \textbf{IC-Defect-14}, including i) multi-cluster head classes, ii) large intra-class diversity, and iii) high inter-class similarity. Because these characteristics have not been presented simultaneously in conventional long-tailed datasets, existing classification methods developed based on standard open datasets cannot well handle real-world highly-imbalanced classification tasks of this kind. 

For example, to highlight the distinctive characteristics of IC defect images, Fig.~\ref{fig:teaser} presents (i) the \textit{t}-SNE visuals and (ii) sample images from a real-world IC defect dataset, namely \textbf{IC-Defect-14}, sourced from the production lines of a global IC fabrication company and detailed in Sec.~\ref{section:datasets}. This figure underscores several critical issues that are often overlooked in the literature. 
First, it is evident in Fig.~\ref{fig:teaser}(a), where defect categories are represented by different colors, that the dataset is highly imbalanced, with some dominant (head) classes (e.g., the normal samples in red) potentially comprising numerous data clusters. This results in what we term the ``many-in-few'' property (i.e., multi-cluster head classes) caused by significant intra-class diversity in head classes. Second, in such a highly imbalanced distribution scenario, most defect classes contain only a small number of IC samples (i.e., the ``few-in-many'' property), making it challenging to learn the class-specific defect features entwined with the class-agnostic information related to the production line domain. Third, the decision margins between minority (tail) defect categories are unclear and insufficiently wide. Consequently, conventional classification models face the challenge of learning appropriate decision boundaries for IC defect classification purposes. 
Fourth, Fig.~\ref{fig:teaser}(b), where samples from the same product line are rendered in the same color, shows that IC defect images within minority classes tend to follow a product-oriented distribution rather than a category-oriented one, posing additional challenges for real-world IC defect classification compared to lab tests. 
Finally, the IC image samples shown in Figs.~\ref{fig:teaser}(c) and~\ref{fig:teaser}(d) further illustrate the complexities of real-world IC defect classification. These figures demonstrate that the same defect type can manifest across chips with different designs, while chips with identical designs may exhibit a variety of defects. This variability contributes to the intricate and overlapping feature representations observed in Figs.~\ref{fig:teaser}(a) and (b), posing significant challenges for deploying deep classification models in AOI systems in production environments.

\begin{table}[!t]
    \caption{Data distribution of \textbf{ID-Defect-14}. The \textbf{ID-Defect-14} dataset contains IC images from 7 product lines, each containing up to 14 defact categories. Image samples in \textbf{ID-Defect-14} were collected automatically by Automated Optical Inspection (AOI) devices in real-world scenario. This dataset exhibits a hybrid highly-imbalanced data distribution in both product-line and defect-category aspects. }
    \centering
    \scriptsize
    \begin{tabular}{|c||c|c|c|c|c|c|c|}
    \hline
        %\multicolumn{8}{|c|}{(A) IC-DEFECT-14 Part-A} \\
         %\hline
        Defect & \multicolumn{7}{c|}{Product Code} \\ \cline{2-8}
        Code & 9KA & 2KA & 64K & 67W & 24W & 9MA & 3QA \\ 
         \hline \hline
         0   & 28689 & 16253 & 4751 & 14167 & 18219 & 46 & 1420 \\
         114 & 3436 & 955 & 934 & 1422 & 680 & 3817 & 162 \\
         129 & 52 & 27 & 13 & 44 & 89 & 0 & 3 \\
         126 & 1127 & 2 & 294 & 0 & 144 & 1089 & 23 \\
         116 & 111 & 108 & 2275 & 197 & 34 & 38 & 20 \\
         132 & 130 & 1335 & 0 & 655 & 0 & 0 & 0 \\
         14  & 63 & 100 & 319 & 204 & 8 & 14 & 3 \\
         111 & 8 & 0 & 439 & 2 & 5 & 0 & 0 \\
         37  & 4 & 0 & 1 & 1 & 4 & 2 & 0 \\
         113 & 3 & 1 & 7 & 5 & 0 & 17 & 86 \\
         158 & 8 & 0 & 11 & 9 & 1 & 26 & 1 \\
         153 & 1 & 0 & 403 & 0 & 6 & 22 & 0 \\
         149 & 0 & 0 & 0 & 0 & 0 & 6066 & 0 \\
         125 & 0 & 0 & 0 & 0 & 0 & 0 & 5716 \\

         \hline 
         
\iffalse         
         \hline
        \multicolumn{8}{|c|}{\textcolor{blue}{(B) IC-DEFECT-14 Extra-Test}} 
         \\
         \hline
        \textcolor{blue}{Defect} & \multicolumn{7}{c|}{\textcolor{blue}{Product Code}} \\ \cline{2-8}
        \textcolor{blue}{Code} & 9KA & 2KA & 64K & 67W & 24W & 9MA & 3QA \\ 
         \hline \hline
         0   & 65 & 301 & 28 & 44 & 411 & 38 & 393 \\
         114 & 313 & 94 & 100 & 155 & 53 & 251 & 67 \\
         129 & 0 & 0 & 3 & 2 & 1 & 1 & 2 \\
         126 & 95 & 0 & 56 & 0 & 27 & 119 & 29 \\
         116 & 10 & 19 & 200 & 45 & 5 & 38 & 2 \\
         132 & 0 & 72 & 0 & 200 & 0 & 0 & 2 \\
         14  & 7 & 12 & 51 & 49 & 2 & 43 & 4 \\
         111 & 8 & 0 & 50 & 0 & 0 & 0 & 0 \\
         37  & 1 & 0 & 0 & 3 & 0 & 0 & 0 \\
         113 & 0 & 0 & 3 & 2 & 1 & 5 & 0 \\
         158 & 1 & 2 & 3 & 0 & 0 & 0 & 1 \\
         153 & 0 & 0 & 6 & 0 & 0 & 5 & 0 \\
         149 & 0 & 0 & 0 & 0 & 0 & 0 & 0 \\
         125 & 0 & 0 & 0 & 0 & 0 & 0 & 0 \\
         \hline
\fi
    \end{tabular}
    \label{tab01:defectdistri}
\end{table}

%\textcolor{blue}{
The data properties of IC-Defect-14 are absent in conventional long-tailed datasets (e.g., iNaturalist-2018~\cite{van2018inaturalist} and ImageNet-LT~\cite{liu2019large}) and remain largely underexplored. This is because existing open datasets are typically curated after collection, leading to the deliberate exclusion of certain data characteristics. As illustrated in Fig.~\ref{fig02:distribution_comparison}, while the curated iNaturalist-2018~\cite{van2018inaturalist} follows an ideal exponential distribution, IC-Defect-14's defect categories are not only extremely imbalanced but also contain data from different production lines in varying proportions. The proportion of data sourced from different production lines within each defect category is detailed in Table~\ref{tab01:defectdistri}, highlighting the fundamental differences between defect classification in the IC manufacturing industry and conventional imbalanced classification techniques. As a result, open datasets like iNaturalist-2018 and ImageNet-LT fail to fully capture the complexities of real-world imbalanced scenarios, making them inadequate representations of naturally occurring data distributions. Consequently, existing highly imbalanced visual recognition techniques, which are developed and validated on standard open long-tailed datasets~\cite{zhang2023deep}, often struggle to effectively handle the unique challenges of IC defect classification. 

%\textcolor{blue}{
To address these challenges, we propose the \textbf{Re}gional \textbf{C}hannel \textbf{A}ttention-based \textbf{M}ulti-\textbf{E}xpert \textbf{Net}work (\textbf{ReCAME-Net}). ReCAME-Net integrates a Regional Channel Attention (RC-Attn) module, metric learning losses, a hard category mining strategy, and a knowledge distillation routine. Extensive experiments demonstrate that our model not only outperforms existing classification networks on the real-world IC-Defect-14 dataset but also maintains competitive performance on standard highly imbalanced datasets, such as ImageNet-LT~\cite{liu2019large} and iNaturalist-2018~\cite{van2018inaturalist}.
%}

Our contributions are summarized as follows. 
\begin{itemize}
    \item To the best of our knowledge, we are among the first to to examine the impact of real-world data characteristics---such as multi-cluster head classes, large intra-class diversity, and high inter-class similarity---on highly imbalanced classification tasks.
    \item We propose a novel method, ReCAME-Net, designed to effectively address the challenges posed by these complex data characteristics, which significantly hinder the performance of existing long-tailed classification approaches.
    \item ReCAME-Net achieves state-of-the-art performance on the highly imbalanced real-world IC image dataset \textbf{IC-Defect-14}, and demonstrates strong competitiveness on widely-used benchmark long-tailed datasets.
    \item Unlike previous state-of-the-art methods that rely on prior dataset-specific information, ReCAME-Net operates without requiring such knowledge, making it highly suitable for practical deployment in IC manufacturing.
\end{itemize}

%% file: TCAD_sec02_review.tex
%In current computer vision researches, imbalanced tasks are frequently treated as long-tailed issues, despite certain distinct differences in properties between real-world highly imbalanced datasets and existing long-tailed datasets. Therefore, we still consider current long-tailed methods as related work.

\subsection{Industrial Classification Missions}
\label{subsec:industrail}
%\textcolor{blue}{
Learning-based data-driven techniques have been widely employed in the field of IC manufacturing to address practical problems, such as automatic defect classification \cite{liu2023machine,xama2023boosting}, wafer failure pattern classification \cite{geng2021wafer,yang2020enhancing}, lithography simulation \cite{GANOPC2018yang,ye2019lithogan,lithonet}, and layout novelty detection~\cite{shao2023keeping}. Most of these practical problems are related to classification techniques, of which primary recent studies seek a solution to highly-imbalanced data classification issues. For example, Alawieh \textit{et al}. proposed using GAN synthesis and data augmentation to assist the representation learning of minority classes (\textit{aka} tail classes) in imbalanced-distributed wafer defect classification problems~\cite{alawieh2020wafer}. This approach was later extended to imbalanced fault classification scenarios by Jiang and Ge~\cite{jiang2020data}. Subsequently, Geng \textit{et al}. combined few-shot~\cite{snell2017prototypical} and self-supervised learning~\cite{chen2020simple} methods to handle minority classes, aiming to address the recognition of mixed types of wafer failure patterns~\cite{geng2021wafer,geng2023mixed}.  
However, these classification techniques developed for semiconductor manufacturing applications usually did not take into account real-world highly-imbalanced data properties like inter-class similarity, intra-class diversity, and multi-cluster head class, resulting from the two common phenomena: \textit{different defect patterns on the same IC design} and \textit{the same defect pattern on different IC designs}, encountered in real IC manufacturing practices. Therefore, we propose a novel classification model to address the challenges posed by these real-world data characteristics.
%}

\subsection{Highly-imbalanced and Long-tailed Data Classification}
\label{section: previous LT method}

Classifying highly-imbalanced data presents a significant challenge. Highly-imbalanced data can be considered a subtype of long-tailed distributed data, wherein the majority of data samples belong to a few head classes, \textit{a.k.a.} majority classes, dominating the dataset, while the remaining tail classes, \textit{a.k.a.} minority classes, share limited amounts of data, resulting in a right-skewed long tail on the data histogram. This severe class imbalance often leads to significantly decreased classification performance in tail classes.

\begin{figure*}[!ht]
    \centering
    \includegraphics[width=0.85\textwidth]{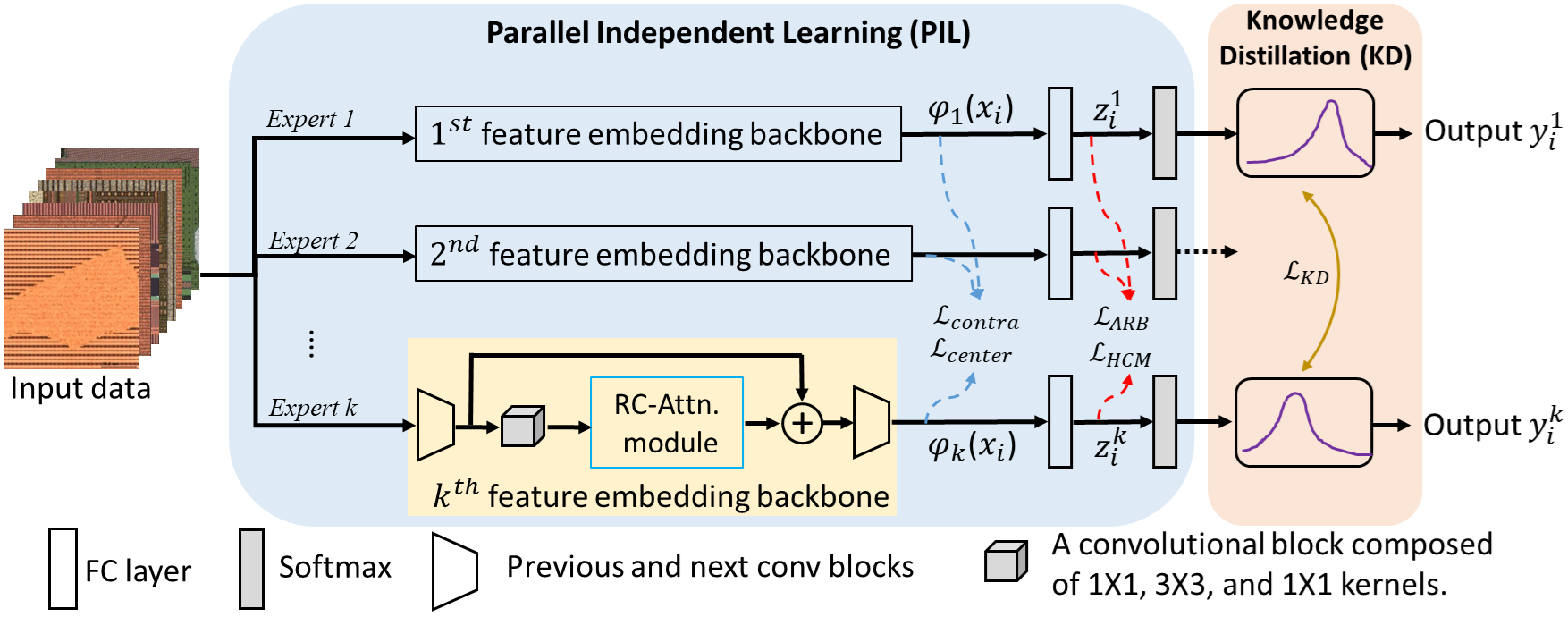}
    %{my_image/Framework.PNG}
    %\captionsetup{font={bf,large}}
    \caption{Architecture of the proposed Regional Channel Attention-based Multi-Expert Network (ReCAME-Net). ReCAME-Net comprises of two stages: Parallel Independent Learning (PIL) and Knowledge Distillation (KD). }
    \label{ReCAME-Net}
\end{figure*}

%\textcolor{blue}{
Existing long-tailed classification techniques can be divided into three major categories: class re-balancing~\cite{cui2019class, huang2016learning,  ren2020balanced, cao2019learning}, decoupled training~\cite{zhong2021improving, zhou2020bbn, kang2019decoupling}, and ensemble learning~\cite{wang2020long,li2022nested, jin2023long, xiang2020learning, zhang2022self, Zhao_2023_ICCV}. Class re-balancing methods include re-sampling~\cite{buda2018systematic} and re-weighting~\cite{cao2019learning}. These techniques aim to modify the decision boundary by rebalancing the contributions per class during training. While class-rebalancing methods can effectively improve the classification accuracy for minority classes by repositioning the decision boundary toward the majority classes, they also sacrifice in the accuracy of majority classes. 
By contrast, decoupled training strategies partition the model training procedure into several phases, each focusing on a specific goal like representation learning or classifier training. For instance, the methods proposed in~\cite{zhou2020bbn,kang2019decoupling} divide the training process into two overlapped phases: the former for representation learning, and the latter for classifier learning. Similarly, Zhong~\textit{et al.}~\cite{zhong2021improving} proposed to initially enhance representation learning with data mixup and subsequently perform classifier learning with a label-aware smoothing strategy. However, these methods were developed based on the presumption that data samples in each class form only one cluster in the feature space, making them unable to tackle the property of multi-cluster head classes in real-world IC defect patterns. 
Recently, ensemble-based strategies have emerged in the form of multi-expert classifiers in the ~\cite{cai2021ace,li2022nested,Zhao_2023_ICCV}. These methods typically employ a parallel-branch design, with each branch specializing in recognizing specific classes. However, most multi-expert systems require prior knowledge of the data distribution for configuring hyperparameters or necessitate picking a relatively balanced data subset for training. Hence, applying techniques of this sort to highly imbalanced real-world data is challenging due to the lack of both prior knowledge of data distribution and a balanced data subset. 
%} 
%
%\textcolor{purple}{indeed simultaneously improve the performance of both head classes and tail classes, but they also come with some limitations and drawbacks. For example, most multi-expert methods assign relatively balanced sub-datasets or data with different distributions to different experts. However, the mutual supervision is insufficient and it is hard to split highly imbalanced datasets into roughly balanced subsets. Additionally, multi-expert networks often lack consideration for feature enhancement and require higher demands for hyper-parameter tuning. }
%
%\textcolor{blue}{
%To address the challenges posed by the real-world highly imbalanced IC defect patterns, 
Thus, we devise a novel multi-expert classification framework to address this problem by incorporating metric learning strategies.
%

%, as will be elaborated in Section~\ref{sec:method}.
%}

%Except for the above drawbacks, based on GCL~\cite{li2022long}, SBCL~\cite{hou2023subclass}, IFL~\cite{tang2022invariant}, and MDCS~\cite{Zhao_2023_ICCV}, we summarize the shortcomings of existing state-of-the-art methods for long-tailed recognition over recent years. These shortcomings include the following points. 1) The re-balancing methods will make the boundary of tail classes that encompass many negative (majority) samples. 2) The computational complexity and difficulty significantly increase when dividing the head class into sub-classes. 3) Although IFL~\cite{tang2022invariant} has raised problems similar to multi-cluster, using center loss alone cannot solve the issue of within class cluster being far away or across other categories. 4) High demand for meticulous adjustments is required in both data augmentation and logit adjustment. As a result, they face challenges in addressing real-world highly imbalanced problems.

\subsection{Attention mechanism}
%\textcolor{blue}{
Since the advocacy of attention mechanisms in~\cite{vaswani2017attention}, attention mechanisms have been widely utilized in recent learning-based approaches to improve feature representability.  %There is an inseparable relationship between the performance of a classification task and the quality of the feature map. Hence, it is crucial to effectively leverage the features extracted by Convolutional Neural Network (CNN). Many studies~\cite{hu2018squeeze, park2018bam, wang2018parameter, woo2018cbam} incorporate attention modules to adjust the importance of spatial and channel domains in the feature map. 
For example, Zhu \textit{et al.} \cite{zhu2021hotspot} devised a hotspot detection method based on an attention module design. Similarly, Shao \textit{et al.} \cite{shao2023keeping} utilized the self-attention mechanism in~\cite{zhang2019self} to enhance feature representation for layout novelty detection. 
The attention mechanism has several extensions to fit different applications. For example, Hu \textit{et al.} constructed SENet~\cite{hu2018squeeze}, which utilizes the squeeze-and-excitation block to derive channel attention for guiding the model on what it is supposed to focus on. Wang \textit{et al.} utilized a spatial attention mechanism to enhance a feature tensor by assessing how the \textit{i}-th position contributes to the \textit{j}-th position within it \cite{wang2018parameter}. Additionally, the Convolutional Block Attention Module (CBAM) in~\cite{park2018bam} integrates spatial attention with channel attention which has been widely used in classifier and object detector designs. However, because CBAM is conceptually a concatenation of a spatial attention module and a channel attention module, the information loss caused by channel reduction during the former channel-attention computation may degrade the performance of the later spatial attention results. 
As a result, we devise a regional-channel attention module to not only address this problem but also make our proposed classifier have a better feature representation for highly-imbalanced IC defect image classification. 
%}

\iffalse
\subsection{Knowledge distillation}
In the beginning, Bucila et al.~\cite{buciluǎ2006model} proposed knowledge distillation and used it for model compression. Later, Hinton et al.~\cite{hinton2015distilling} extended the applicability of knowledge distillation, making it more suitable for classification tasks. Subsequent researches also adopted the teacher-student learning (offline distillation) framework proposed by Hinton et al.~\cite{hinton2015distilling}. For instance, ~\cite{furlanello2018born, mirzadeh2020improved} consider the output logits generated by a large teacher model with rich information as the target, and then distill knowledge to a shallow student model through soft targets. In addition, in recent years, with the rise of multi-expert models, more and more studies have been using online distillation~\cite{guo2020online, yang2023online, anil2018large, zhang2018deep}. Compared to offline distillation, in online distillation, both the teacher model and the student model are trained together and collaborate with each other.  In other words, in online distillation, each model not only guides other models but also distills knowledge from other models. Therefore, online distillation is efficient and complementary in aligning the features or output logits of multiple experts.
\fi

%% file: TCAD_sec03_method.tex
\subsection{Overview}
%\textcolor{blue}{
To address the challenges posed by the highly imbalanced real-world IC defect image dataset \textbf{IC-Defect-14}, we devise the Regional Channel Attention-based Multi-Expert Network (\textbf{ReCAME-Net}), which integrates metric learning and attention mechanisms within a multi-expert framework. Existing long-tailed classification methods are typically designed without considering the presence of multi-cluster head classes, a phenomenon commonly observed in real-world industrial defect classification. In contrast, ReCAME-Net incorporates two metric learning losses to condense intra-class clusters, aligning with the design assumptions of existing long-tailed classification strategies. Specifically, the ARB loss~\cite{xie2023neural}, which is designed to mitigate the minority collapse problem, is theoretically derived under the assumption that each class consists of a single cluster, a limitation that ReCAME-Net effectively addresses~\cite{dang2023neural, yang2022inducing}.
%}

%\textcolor{blue}{
Fig.~\ref{ReCAME-Net} depicts the architecture of ReCAME-Net. The model adopts a multi-expert framework, where multiple classifier branches are trained, each integrating a metric learning strategy, an ARB loss for minority collapse mitigation, a hard-category mining routine, and an attention-based feature enhancement module. More specifically, the metric learning strategy is designed to merge intra-class clusters, ensuring that the assumptions required by the ARB loss for minority collapse mitigation hold. Additionally, the hard-category mining loss plays a crucial role in maintaining well-defined decision boundaries between challenging classes and their neighboring classes in the feature space, enhancing classification robustness in real-world IC defect datasets.  To ensure a coherent and reliable classification output across multiple expert branches, ReCAME-Net incorporates a knowledge distillation process, which aligns predictions from all classifier branches while suppressing potential misclassifications. This collaborative learning process enhances the model’s ability to generalize across imbalanced defect distributions, making ReCAME-Net a powerful solution for industrial defect classification in high-yield manufacturing scenarios.  
%}

Note that the number of branches does not need to match the number of defect categories, as branches are not required to specialize in any specific class. Each branch of ReCAME-Net functions as an independent classifier, initialized with its own parameters and trained to converge toward an optimal solution. This enables classifiers to capture complementary strengths, with some learning certain categories more effectively than others. The KL-divergence and classification loss then guide all classifiers to align while converging toward a solution that maximizes overall performance.

\subsection{Parallel independent learning (PIL)}
\label{section3.2}

\begin{figure}[!t]
    \includegraphics[width=0.48\textwidth]{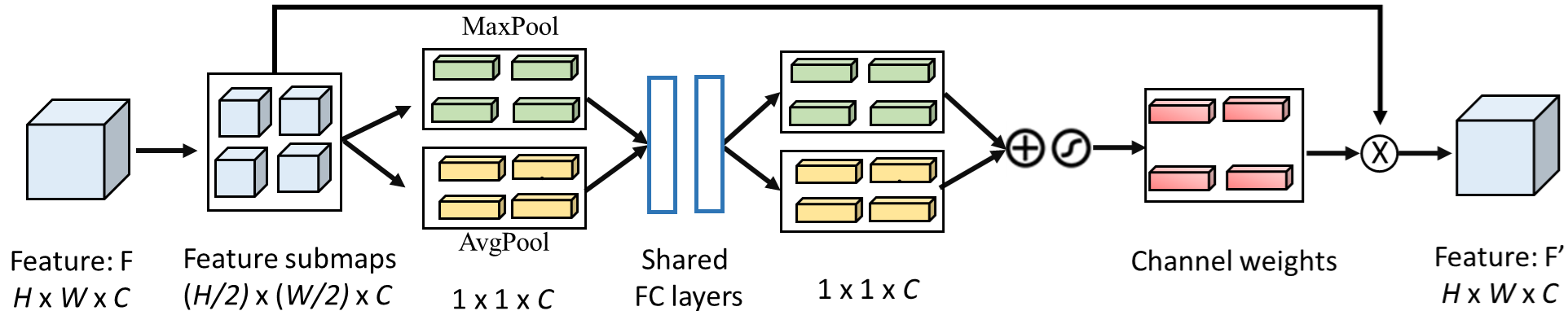}
    %{my_image/RC Attn.PNG}
    %\captionsetup{font={bf,large}}
    \caption{Architecture of the proposed RC-Attn module. }
    \label{RC-Attn}
\end{figure}

%\textcolor{blue}{
In our design, an expert branch in the PIL stage is implemented as a ResNet-50 with a regional-channel-attention (RC-Attn) module for feature enhancement. 
The structure of the RC-Attn module is shown in Fig.~\ref{RC-Attn}. After dividing a feature tensor of size $H \times W \times C$ into four $\frac{1}{2} H \times \frac{1}{2} W \times C$ sub-tensors, each sub-tensor is then independently refined using the channel attention mechanism from~\cite{woo2018cbam} to enhance its feature representation.  This sub-tensor splitting strategy preserves spatial correlations within localized regions, ensuring that critical spatial information is retained throughout the channel attention computation process. This design further strengthens the feature representability for classification tasks.

\begin{figure}[!t]
    \includegraphics[width=0.48\textwidth]{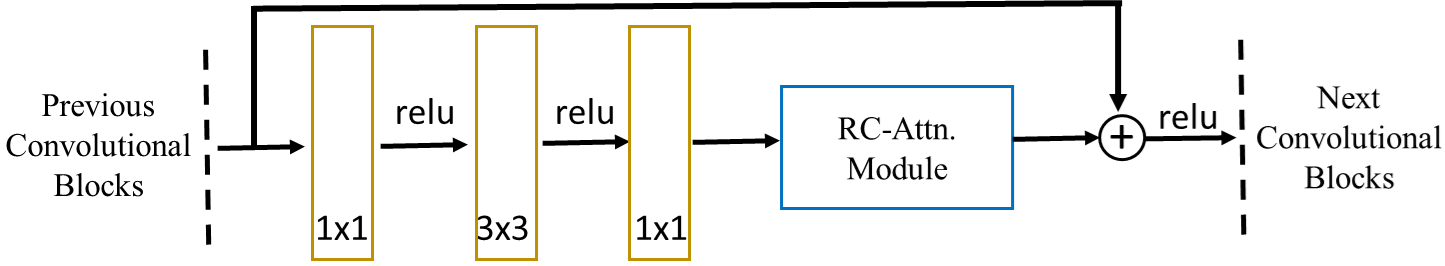}
    %{my_image/RC with Resnet.PNG}
    %\captionsetup{font={bf,large}}
    \caption{Detailed architectural design of RC-Attn module. %Exact position of RC-Attn module in ResBlock for ResNet~\cite{he2016deep}. We implemented our module on the convolutional output of each block.
    }
    \label{RC with Resnet}
\end{figure}

The integration of the RC-Attn module with ResNet-50~\cite{he2016deep} is depicted in Fig.~\ref{RC with Resnet}.   
In the PIL stage, each expert branch functions as an independent classification network, trained separately to optimize performance. The training process incorporates key strategies, including mitigating minority collapse~\cite{dang2023neural, xie2023neural, yang2022inducing}, metric learning for mitigating intra-class diversity, feature enhancement via attention mechanisms, and hard category mining~\cite{li2022nested} to refine decision boundaries.

\subsubsection{Attraction-Repulsion-Balanced (ARB) Loss }

%\textcolor{blue}
%{
Recent studies have shown that when training a classification network using the standard cross-entropy loss, classes with large sample sizes can push decision boundaries outward, causing minority class decision boundaries to collapse---a phenomenon known as minority collapse~\cite{dang2023neural, xie2023neural, yang2022inducing}. These findings provide a mathematical explanation for the challenge of training classifiers capable of effectively recognizing tail-class samples in highly imbalanced datasets.

To address this issue, Xie \textit{et al}. introduced the Attraction-Repulsion-Balanced (ARB) loss~\cite{xie2023neural}, which was derived through theoretical analysis. They further demonstrated that the ARB loss is mathematically equivalent to the balanced meta-softmax loss proposed in~\cite{ren2020balanced}, a method specifically designed for handling class imbalance in classification tasks.
Therefore, instead of using cross-entropy loss, we adopt the ARB loss as the primary loss function for ReCAME-Net, ensuring more balanced decision boundaries for highly imbalanced IC defect classification. The ARB loss is formulated as:
\begin{equation}
    \mathcal{L}_\mathrm{ARB} = - \sum_{j}\log\bigg( \frac{e^{z_{j}}}{\sum_{k=1}^C \frac{n_k}{n_j} e^{z_{k}} } \bigg),
    \label{eq:ARB}
\end{equation}
where $z_j$ is the input of the softmax layer, and $n_j$ denotes the number of samples in the $j$-th class. By incorporating the ARB loss, our model learns more appropriate decision boundaries, effectively mitigating the minority collapse issue and improving classification performance in real-world imbalanced datasets. 
%}

\iffalse
The cross-entropy loss is the most common loss function used in classification tasks, but it is not suitable for imbalanced distributions. When directly apply cross-entropy loss to long-tailed dataset, it tends to produce results that are dominated by the head classes. Moreover, this phenomenon becomes more pronounced as the imbalance ratio increases, the imbalance ratio is defined as $n_{max}$/$n_{min}$, where $n_{max}$ and $n_{min}$ represent the maximum and minimum class sizes, respectively.

Because ~\cite{dang2023neural, xie2023neural, yang2022inducing} indicate that a class with a greater number of samples could potentially extend the classification boundary farther from its prototype, leading to the collapse of minority classes. Thus, recent developments in loss functions based on cross-entropy, e.g.,  ~\cite{ren2020balanced , xie2023neural} have incorporated margin adjustments to promote competition among classes. Based on previous studies, we utilize the ARB-Loss in ~\cite{xie2023neural} instead of the original cross-entropy loss. That is
\begin{equation}
    \mathcal{L}_\mathrm{ARB} = - \sum_{k,j}\log\bigg(\frac{\exp(z^k_{i,j})}{\sum_{l=1}^C \frac{n_l}{n_j}\exp(z^k_{i,l})} \bigg)
    \mbox{,}
    \label{eq:ARB}
\end{equation}
where $z^k_{i,j}$ is the class-j output of $k$-th expert for $i$-th sample, $n_j$ denotes the cardinality of $j$-th class in the training set, and $C$ is the number of classes. In this way, the model can generate results that are more suitable for highly imbalanced datasets, and prevent the collapse of minority classes.
\fi

\subsubsection{Metric Learning Loss}

%\textcolor{blue}
%{
Since real-world IC defect image properties do not fully align with the assumptions underlying the derivation of the ARB loss, we introduce two metric learning loss terms to merge intra-class clusters and widen the margin between similar classes. These enhancements enable the effective application of the ARB loss in real-world scenarios.

To centralize intra-class clusters and improve feature compactness, we define the following centering loss:
\begin{equation}    
\mathcal{L}_{\mathrm{cen}} = 
    %\mathbb{E} \{ 
    \mathbb{E}[ \delta_{ij}\cdot \| \phi_k(x_j) - \phi_k(x_i) \|_2 ],
    \label{eq:center}
\end{equation}
where $\| \cdot \|_2$ represents the $L_2$ norm, and $x_i$ and $\phi_k(x_i)$ denote the $i$-th training sample and its feature embedding from the $k$-th backbone, respectively. The expectation operator $\mathbb{E}[ \cdot ]$ is applied per training batch. This loss encourages intra-class samples to be drawn closer together in feature space, improving class cohesion. 

To ensure sufficient separation between similar classes, we introduce a contrastive loss, defined as: 
\begin{equation}
    \mathcal{L}_{\mathrm{cont}} = \mathbb{E}[ \delta_{ij} \cdot d_{ij}
    + (1-\delta_{ij}) \max(0, d_\mathrm{MG} - d_{ij})],
    \label{eq:interloss}
\end{equation}
where $d_{ij} = \| \phi(x_i) - \phi(x_j) \|_2$ represents the $L_2$ distance between feature embeddings, and $\delta_{ij}$ is an indicator function where $\delta_{ij}=1$ if $x_i$ and $x_j$ belong to the same class, and $\delta_{ij}=0$ otherwise. This loss enforces a minimum inter-class margin, ensuring that feature spaces of different classes remain well separated. The margin is set to a default value of $d_\mathrm{MG}=1.0$.

It is worth noting that (\ref{eq:center}) is mathematically equivalent to the first term in (\ref{eq:interloss}). However, each loss term serves a distinct role in feature representation learning—one focusing on intra-class compactness, while the other enforces inter-class separation.
\subsubsection{Hard Category Mining (HCM) Loss}
\label{subsubsec:HCMloss}

%\textcolor{blue}
%{
In real-world scenarios, certain data categories remain particularly challenging to classify accurately. To address this, we adopt the Hard Category Mining (HCM) strategy proposed in~\cite{li2022nested}, which directs the model's attention toward these difficult-to-classify categories.
The HCM loss, $\mathcal{L}_{\mathrm{HCM}}$, computes cross-entropy using the conditional probability of the target hard classes. It operates based on a hard category set $\Psi$, which consists of classes that are prone to misclassification. Specifically, these hard classes are identified as the union of the target ground-truth class and the Top-$N$ predicted classes with the highest probabilities (i.e., those whose features closely resemble the ground-truth class). By penalizing misclassifications within this subset of confusing classes, the HCM loss helps the model focus on distinguishing between closely related categories. The HCM loss is applied on a branch-wise basis and is defined as:
\begin{equation}
    \mathcal{L}_\mathrm{HCM} = -  \log \bigg( \frac{\exp(z^*)}{\sum_{l\in \Psi} \frac{n_l}{n^*} \exp(z_l)} \bigg),
\end{equation}
where $z^*$ and $n^*$ represent the logit and sample count of the target class, respectively. $z_l$ and $n_l$ are defined similarly for each class 
$l$ in $\Psi$. 

Beyond improving classification accuracy for hard-to-distinguish categories, the HCM loss also helps prevent tail classes from becoming persistently misclassified hard classes. By penalizing instances of minority collapse, it ensures that rare defect categories remain well-represented in the learned feature space. 
%}

\iffalse
In classification tasks, some non-target classes are more likely to be misclassified as target. This phenomenon is especially noticeable in the head classes of long-tailed datasets. For instance, minority class samples are often easily misclassified as head classes. These non-target classes with high confidence are referred to as hard categories.
Based on above reason, we utilize the hard category mining (HCM) strategy introduced in ~\cite{li2022nested} to alleviate the classification uncertainty. For $i$-th training sample $x_i$ and $k$-th expert, $\Psi_i^k$ defined as the union of the target class and the classes with the Top-N prediction probabilities. The hard category mining loss $\mathcal{L}_\mathrm{HCM}$ also follows the concept of ARB-loss, which is shown as:
\begin{equation}
    \Psi_i^k = TopN\big\lbrace{z^k_{i,j}|j\neq y_i \big\rbrace} \cup \big\lbrace{z^k_{i,y}\big\rbrace}
    \mbox{, and}
    \label{eq: Psi}
\end{equation}
\begin{equation}
    \mathcal{L}_\mathrm{HCM} = - \sum_{k,j}\log\bigg(\frac{\exp(z^k_{i,j})}{\sum_{z^k_l\in\Psi_i^k} \frac{n_l}{n_j}\exp(z^k_{i,l})} | z^k_{i,j} \in \Psi_i^k \bigg)
    \mbox{,}
    \label{eq:HCM}
\end{equation}
Note that the hard category set $\Psi_i^k$ is determined by the original softmax function instead of cardinality, as the inference result also depends on the conventional softmax.
\fi

\begin{table*}[!t]
    \caption{Details of three imbalanced datasets used in this study}
    \footnotesize
    \centering
%    \small
    \begin{tabular}{|c|c|c|c|c|c|c|}
    \hline 
    \textbf{Dataset} & \textbf{\# of} & \textbf{Imbalance} & \textbf{Training} & \textbf{Validation} & \textbf{Testing} & \textbf{Testing}  \\
    & \textbf{Classes} & \textbf{Ratio} & \textbf{Samples} & \textbf{Samples} & \textbf{Samples} & \textbf{Set} \\
      \hline
      \textbf{IC-Defect-14} & 14 & 50127/7 & 69,789 & - & 46,538 & imbalanced\\
      \hline
      \textbf{ImageNet-LT~\cite{liu2019large}} & 1,000 & 1280/5 & 115,846 & 20,000 & 50,000 & balacned\\
      \hline
      \textbf{iNaturalist 2018~\cite{van2018inaturalist}} & 8,142 & 1000/2 & 437,513 & 24,426 & - & balacned\\
      \hline
    \end{tabular}
    %\captionsetup{font={bf,large}}
    \label{tab:dataset Comparison}
\end{table*}

\subsection{Knowledge Distillation}
\label{section3.3}
%\textcolor{blue}
%{
The concept of knowledge distillation was first pioneered in~\cite{buciluǎ2006model}, with Hinton \textit{et al}. later introducing the knowledge distillation loss  to minimize discrepancies between the output distributions of multiple classification networks' final softmax layers. Inspired by Hinton’s work, recent multi-expert classifiers designed for long-tailed or highly imbalanced data~\cite{li2022nested, zhang2023balanced, park2023mutual} have integrated knowledge distillation routines to harmonize expert outputs and achieve a robust final consensus.
In ReCAME-Net, we employ the knowledge distillation loss as proposed by Li \textit{et al}. in their NCL classifier~\cite{li2022nested}, which consists of two components:  
\begin{equation}
    \mathcal{L}_\mathrm{KD} =  \mathcal{L}^\mathrm{all}_\mathrm{KD} + \mathcal{L}^\mathrm{hard}_\mathrm{KD},
    \label{eq:KD}
\end{equation}
where $\mathcal{L}^\mathrm{all}_\mathrm{KD}$  manages the distillation across all classes and experts, and  $\mathcal{L}^\mathrm{hard}_\mathrm{KD}$ focuses on aligning predictions for hard categories. 

Both loss terms are based on Kullback-Leibler (KL) divergence, defined as follows:
\begin{equation}
  \mathcal{L}^\mathrm{all}_\mathrm{KD} = \mathbb{E}[\mathrm{KL}\big(P_{z^k}|P_{z^l})],
    \label{eq:KD_all}
\end{equation}
and
\begin{equation}
  \mathcal{L}^\mathrm{hard}_\mathrm{KD} = \mathbb{E}[\mathrm{KL}\big(P_{z^k}^*|P_{z^l}^*)],
    \label{eq:KD_hard}
\end{equation}
where $P_{z^k}$ represents the per-batch distribution of classification probability vectors predicted by the $k$-th expert (branch), while $P^*_{z^k}$ denotes the corresponding distribution for hard categories.

Finally, the $j$-th entry of the $i$-th sample's probability vector, predicted by the $k$-th branch, is derived from the $k$-th branch's final softmax layer by using %taking $\textbf{z}^k_{i,j}$ as input
\begin{equation}
    \mathbf{P}_{z^k}(j ; \mathbf{x}_i, \theta_k) = \frac{e^{z^k_{i,j}}}{\sum_{l} \frac{n_l}{n_j} e^{z^k_{i,l}}},  
 \label{eq:classprob}
\end{equation}
where $\theta_k$ denotes the parameter of the $k$-th classifier, $n_j$ and $n_l$ represent the sample counts of the target and other classes, respectively. 

By leveraging the balanced meta-softmax function~\cite{ren2020balanced} in (\ref{eq:classprob}), ReCAME-Net ensures that class imbalance is accounted for in the probability estimation, improving both inter-expert consensus and performance on imbalanced IC defect classification tasks.

\subsection{Total Loss and Loss Weight Selection}

%\textcolor{blue}

As a result, the total loss of ReCAME-Net is the sum of all loss terms used in the PIL and KD stages. That is, 
\begin{eqnarray}
    \mathcal{L}_\mathrm{Total} &=& \mathcal{L}_\mathrm{ARB} + \mathcal{L}_\mathrm{HCM} + w_1\mathcal{L}_{\mathrm{cont}} +  w_2 \mathcal{L}_{\mathrm{cen}} \nonumber\\
    && + \alpha (\mathcal{L}^\mathrm{all}_\mathrm{KD} + \mathcal{L}^\mathrm{hard}_\mathrm{KD}),
    \label{eq:totalloss}
\end{eqnarray}
where $w_1=0.05$, $w_2=0.000625$, and $\alpha=0.6$ are used for all benchmarks.

The loss weights in (\ref{eq:totalloss}) were determined through a step-wise development process. Starting from a baseline model trained with only %the ARB loss 
$\mathcal{L}_\mathrm{ARB}$, we gradually integrated %the contrastive loss  
$\mathcal{L}_\mathrm{cont}$,  %centering loss 
$\mathcal{L}_\mathrm{cen}$, %hard category mining loss 
$\mathcal{L}_\mathrm{HCM}$, and %knowledge distillation losses 
$(\mathcal{L}^\mathrm{all}_\mathrm{KD}+\mathcal{L}^\mathrm{hard}_\mathrm{KD})$ in sequence, tuning only the weight of the newly introduced loss term at each step via $K$-fold cross-validation.
The weighting coefficients %used in (\ref{eq:totalloss}) 
were carefully validated through multiple rounds of experimentation on both internal and public datasets. Finally, note that the HCM loss is weighted equally to the ARB loss ($1.0$), owing to its consistently stable effectiveness across all experiments.

\iffalse
Finally, the overall training loss of ReCAME-Net is formulated as:
\begin{equation}
  \mathcal{L}_\mathrm{Total} = \mathcal{L}_\mathrm{PIL} + \alpha \mathcal{L}_\mathrm{KD}
    \mbox{,}
    \label{eq:probability}
\end{equation}
where $\mathcal{L}_\mathrm{PIL}$ is the total loss of the parallel independent learning, $\mathcal{L}_\mathrm{KD}$ as previously described in section~\ref{section3.3}, and $\alpha$ represent a weighting factor for cooperation among each expert (branch). As mentioned in section~\ref{section3.2}, the parallel independent learning loss $\mathcal{L}_\mathrm{PIL}$ is composed by four component, i) a attraction-repulsion-balanced loss $\mathcal{L}_\mathrm{ARB}$  , ii) a hard category mining loss $\mathcal{L}_\mathrm{HCM}$, and iii) two metric learning losses $\mathcal{L}_\mathrm{contra}$  and $\mathcal{L}_\mathrm{center}$  , so we can summarize as:
\begin{equation}
    \mathcal{L}_\mathrm{PIL} =  \mathcal{L}_\mathrm{ARB} + \mathcal{L}_\mathrm{HCM} + w_1\mathcal{L}_{\mathrm{contra}} +  w_2 \mathcal{L}_{\mathrm{center}}
    \mbox{,}
    \label{eq:KD}
\end{equation}
where $w_1=0.05$ and $w_2=0.000625$ are weighting factors that we initially set up.
\fi

%% file: TCAD_sec04_experiment.tex
\subsection{Datasets}
\label{section:datasets}

%\textcolor{blue}
All our experiments were conducted on a real-world highly-imbalanced IC defect image dataset, IC-Defect-14, provided by the output quality assessment department of a global IC manufacturing company. IC-Defect-14 comprises 14 categories of IC chip image data, where 1 class consists of all normal regular images from multiple product lines, and the remaining 13 classes contain defect images. The training set of this dataset contains 69,789 samples\footnote{We divide the IC-Defect-14 training set into five subsets and apply a leave-one-out cross-validation strategy, where the model is trained on four subsets and evaluated on the remaining one in each iteration.}, while the testing set consists of 46,538 images. Notably, the largest head class contains 50,127 images, whereas the smallest tail class contains only 7 image samples, resulting in a high imbalance ratio of 50,127/7 = 7,161. Additionally, as illustrated in Fig.~\ref{fig:teaser}, the dataset features complex background appearances due to the diverse designs of different product lines, posing a significant challenge for developing a classification network for this real industrial dataset.

Moreover, in order to assess the generalizability of our ReCAME-Net, we conducted two additional sets of experiments on two classical datasets for long-tailed classification, namely ImageNet-LT~\cite{liu2019large} and iNaturalist-2018~\cite{van2018inaturalist}. ImageNet-LT is a subset sampled from the well-known ImageNet dataset~\cite{deng2009imagenet} following the Pareto distribution with a Pareto index of 6, comprising 115.8K images categorized into 1,000 distinct categories. Additionally, iNaturalist-2018 is a large-scale photo dataset where images, uploaded by global users, were selected artificially by iNaturalist superusers for species classification. iNaturalist-2018 consists of 437.5K training images and 24.4K validation images across 8,142 species categories. Finally, Table~\ref{tab:dataset Comparison} lists the details of the three datasets used in this study.

%\newpage
\subsection{Configuration and Implementation Details}

%\textcolor{blue}
{
For the IC-Defect-14 dataset, we employed ResNet-50 as the feature extraction backbone and used SGD with a momentum of 0.9 as the optimizer. The learning rate was scheduled using cosine decay, starting with an initial learning rate of 0.1 and gradually reducing to a minimum learning rate of 0.0001. Our ReCAME-Net was trained for 150 epochs without applying any data augmentation strategies. To enhance the adaptability of ReCAME-Net in highly imbalanced classification tasks, we replaced the conventional linear classification head with the cosine classification head proposed in~\cite{zhang2021distribution}.  For the Top-$N$ setting used in the hard-category-mining loss, as described in Sec.~\ref{subsubsec:HCMloss}, we set $N$ to be 30\% of the total number of categories. For instance, in IC-Defect-14, which has 15 categories, we determined $N$ as $N=\mathrm{floor}( 15\times0.3)=4$, selecting the four most challenging categories for loss calculation. All experiments were conducted on a computer equipped with NVIDIA P100 GPUs }

%\textcolor{blue}
{
For nature image datasets like ImageNet-LT and iNaturalist-2018, the total number of training epochs was set to be 200 with an initial learning rate of 0.2, and the network implementation and other training configurations are identical to those used for the IC-Defect-14 dataset. To improve model generalization, we adopted the data augmentation strategy suggested in \cite{cubuk2020randaugment} for experiments on these datasets. Additionally, we followed the training schedule outlined in \cite{cui2021parametric}. All experiments of this part were conducted using a machine equipped with 8 NVIDIA Tesla V100 GPUs. 
}

\iffalse
For the two open datasets: ImageNet-LT~\cite{deng2009imagenet} and iNaturalist 2018~\cite{van2018inaturalist}, we used ResNet-50 as our backbone, SGD with a momentum of 0.9 as optimizer, and utilized a learning rate schedule based on cosine decay, with an initial learning rate=0.2 and a minimum=0.0001. Moreover, to address long-tailed recognition tasks more effectively, we adopted the cosine classifier proposed by~\cite{zhang2021distribution} instead of a linear classifier. The other training strategies followed~\cite{cui2021parametric} and~\cite{cubuk2020randaugment}. All experiments on both datasets were conducted by 8 NVIDIA Tesla V100 GPUs.

For experiments on UMC dataset, we used ResNet-50 as a backbone. The training epoch is 150 and the initial learning rate is 0.01. The remaining settings and training strategy are same as that mentioned for the open datasets, but we did not employ any data augmentation techniques. All models for highly imbalanced dataset were trained on NVIDIA P100 GPUs.

In all experiments, for the Top-$N$ of the hard category mining loss $\mathcal{L}_\mathrm{HCM}$, we used 30\% of the total number of categories, rounded down to the nearest integer. For example, in the case of ImageNet-LT~\cite{liu2019large}, where the total number of categories is 1000, the $TopN$ would be 300.
\fi

%%==================================
\begin{table*}[!t]
    \caption{Performance comparison on (A) IC-Defect-14's blind testing subset, (B) iNaturalist-2018~\cite{van2018inaturalist}, (C) ImageNet-LT~\cite{liu2019large} }
    \centering
    %\Large
    \footnotesize
    \begin{tabular}{|l|c|c|c|c|c||c|c|c|c||c|}
    \hline 
     & \multicolumn{5}{c||}{(A) IC-Defect-14} & \multicolumn{4}{c||}{(B) iNaturalist-2018  } & (C)~\cite{liu2019large}\\ 
     \cline{2-11}
     & Head & Many & Medium & Few & Avg. & Many & Medium & Few & Avg. & Avg.\\
     Method  & ($>10^4$) & ($10^3$--$10^4$) & ($10^2$--$10^3$) & ($\leq10^2$) &  & ($>10^2$) & ($20$--$10^2$) & ($<20$) &  & \\
     \hline
     ResNet \cite{he2016deep} & \textit{98.72} & 77.33 & 23.77 & 20.78 & 91.76 & 72.2 & 63.0 & 57.2 & 61.7 & 41.6\\
     ResNet w/ $\beta$ & \textit{99.09} & 82.89 & 36.75 & 7.79 & 92.33 & 53.4 & 54.8 & 53.2 & 54.0 & 33.2\\ 
     ResNet w/ focal loss & 97.35 & 82.17 & 40.98 & 18.18 & 93.69 & -- & --- & -- & 61.1 & 42.0 \\ \hline
     RIDE (2021)~\cite{wang2020long} & \textit{99.71} & 74.17 & 5.19 & 0 & 91.28 & 70.2 & 72.2 & 72.7 & 72.2 & 54.9 \\
     TSC (2022)~\cite{li2022targeted} & 95.38 & 52.28 & 13.80 & 23.38 & 82.59 & 72.6 & 70.6 & 67.8 & 69.7 & --\\
     FBL (2022)~\cite{li2022feature} & 96.55 & 70.84 & 40.16 & 0 & 88.50 & -- & -- & -- & 69.9 & 52.4\\
     NCL (2022)~\cite{li2022nested} & 98.33 & 82.24 & 51.23 & 33.77 & 93.23 & 72.7 & 75.6 & 74.5 & 74.9 & 59.5\\
     % MDCS~\cite{Zhao_2023_ICCV} & - & 98.40 & \textcolor{red}{86.00} & 54.64 & \textcolor{blue}{36.36} & \textcolor{red}{94.33} \\
     SBCL (2023)\cite{hou2023subclass} & 43.43 & 56.59 & 34.43& 25.97 & 46.75 & 73.3 & 71.9 & 68.6 & 70.8 & 53.4 \\
     SHIKE (2023)~\cite{jin2023long} & \textit{99.39} & 83.12 & 29.78 & 3.90 & 93.84 & -- & -- & -- & 75.4 & 59.7 \\
     %SHIKE (2023) & 86.80 & 68.88 & 19.92 & 0 & 70.03 & -- & -- & -- & 75.4 & 59.7 \\
     MDCS (2023)~\cite{Zhao_2023_ICCV} & 96.04 & 78.07 & 54.23 & \textit{54.55} & 90.58 & 76.5 & 75.5 & 75.2 & 75.6 & 60.7\\ %育丞補的100epoch版
%     MDCS~\cite{Zhao_2023_ICCV} & \textcolor{blue}{99.40} & \textcolor{blue}{86.30} & 49.60 & 18.18 & \textcolor{blue}{95.01} \\ % 君豪實驗版最佳數值，不要刪
    BalPoE (2023)~\cite{aimar2023balanced} & 82.95 & \textit{85.77} & \textit{61.48} & 36.36 & 83.30 & -- & -- & -- & 75.0 & 59.8\\
     
     \hline
     Wafer2Spike~\cite{mishra2024wafer2spike} & 97.17 & 75.28 & 48.16 & 1.18 & 89.86 & -- & -- & -- & -- & -- \\
     %\hline
     WaferSegClassNet~\cite{nag2022wafersegclassnet} & 97.90 & 77.21 & 34.07 & 0 & 90.40 & -- & -- & -- & -- & --\\
     
     \hline
     \textbf{Ours} & \textbf{98.40} & \textbf{83.77} & \textbf{60.38} & \textbf{40.26} & \textbf{93.83} & \textbf{74.9} & \textbf{75.1} & \textbf{73.1} & \textbf{74.3} & \textbf{58.5}\\
%     Ours (w/o RC-Attn) & 98.45 & 83.30 & 52.46 & 31.17 & 93.61 \\
%     Ours (Spat. Attn) & 98.54 & 82.66 & 57.38 & {40.26} & 93.59 \\
%     Ours (Chan. Attn) & 98.65 & 83.43 & {61.61} & {35.06} & {93.94} \\
%    \hline
%    \hline
%     \textcolor{red}{Yu-Cheng} & \textcolor{red}{99.39} & \textcolor{red}{85.19} & \textcolor{red}{61.61} & \textcolor{red}{30.06} & \textcolor{red}{94.44} & -- & -- & -- & -- & --\\
     \hline
     \hline
     \# of Samples & 33,418 & 12,311 & 732 & 77 & 46,538 & 2,526 & 12,228 & 9,672 & 24,426 & 50k\\
     \hline     
     
    \end{tabular}
    %\captionsetup{font={bf,Large}}   
    \label{tab:umc_inat_imagenet}
\end{table*}

\begin{table*}[!t]
    \caption{Per-category accuracy (\%) comparison on IC-Defect-14 testing set %\textcolor{blue}{and IC-Defect-14EX}
    }
    \centering
    %\Large
    \scriptsize
    \begin{tabular}{|l|c|c|c|c|c|c|c|c|c|c|c|c|c|c|c|}
    \hline 

    %\multicolumn{16}{|c|}{\textcolor{blue}{(A) Per-Defect Accuracy (\%) on IC-Defect-14's Testing Set}}\\ \hline
    
     Method & \multicolumn{14}{c|}{Defect Codes } & Whole \\ \cline{2-15}
     & 0 & 114 & 149 & 125 & 116 & 126 & 132 & 14 & 111 & 153 & 129 & 113 & 158 & 37 & Subset \\
     \hline
     ResNet w/ $\beta$ & 99.09 & 79.73 & 97.73 & 99.96 & 67.95 & 51.87 & 70.17 & 24.21 & 91.21 & 15.61 & 7.61 & 12.50 & 0 & 0 & 92.33 \\
     ResNet w/ focal loss & 97.35 & 75.50 & 98.27 & 99.87 & 72.35 & 56.53 & 69.58 & 31.23 & 90.66 & 24.28 & 4.35 & 29.17 & 0 & 0 & 93.69 \\
     \hline
     RIDE~\cite{wang2020long} & 99.71 & 78.06 & 97.32 & 82.64 & 57.99 & 31.81 & 38.92 & 6.67 & 0 & 10.98 & 0 & 0 & 0 & 0 & 91.28 \\
     TSC~\cite{li2022targeted} & 95.38 & 37.98 & 94.81 & 77.92 & 22.53 & 4.57 & 37.74 & 2.11 & 51.10 & 1.16 & 0 & 37.5 & 0 & 0 & 82.59\\
     FBL~\cite{li2022feature} & 96.55 & 62.81 & 95.10 & 95.98 & 46.95 & 45.62 & 40.09 & 12.98 & 81.32 & 63.01 & 0 & 0 & 0 & 0 & 88.50\\
     NCL~\cite{li2022nested} & 98.33 & 72.30 & 98.48 & 99.87 & 68.04 & 67.44 & 79.01 & 39.30 & 96.15 & 31.79 & 35.87 & 50.00 & 4.35 & 16.67 & 93.23 \\
     % MDCS~\cite{Zhao_2023_ICCV} & - & 98.40 & \textcolor{red}{86.00} & 54.64 & \textcolor{blue}{36.36} & \textcolor{red}{94.33} \\
     SBCL\cite{hou2023subclass} & 43.43 & 31.34 & 94.03 & 91.74 & 24.78 & 38.90 & 54.72 & 21.05 & 58.24 & 36.42 & 25.00 & 37.50 & 0 & 33.33 & 46.75\\
     
     SHIKE~\cite{jin2023long} & 99.39 & 78.44 & 98.07 & 99.83 & 73.79 & 50.56 & 73.94 & 11.93 & 93.96 & 6.94 & 1.09 & 6.25 & 0 & 0 & 93.84\\
     %\textcolor{red}{SHIKE (re)}~\cite{jin2023long} & 99.59 & 78.70 & 98.39 & 99.83 & 71.90 & 47.76 & 65.33 & 11.93 & 88.46 & 6.94 & 1.09 & 2.08 & 0 & 0 & 93.73\\
     MDCS~\cite{Zhao_2023_ICCV} & 96.04 & 62.44 & 98.76 & 99.87 & 67.24 & 60.17 & 81.01 & 41.05 & 97.25 & 34.68 & 46.74 & 87.50 & 0 & 0 & 90.58\\ %育丞補的100epoch版
%     MDCS~\cite{Zhao_2023_ICCV} & \textcolor{blue}{99.40} & \textcolor{blue}{86.30} & 49.60 & 18.18 & \textcolor{blue}{95.01} \\ % 君豪實驗版
    BalPoE~\cite{aimar2023balanced} & 82.95 & 78.39 & 98.52 & 100 & 72.53 & 70.15 & 87.74 & 59.65 & 96.15 & 37.57 & 43.48 & 50.00 & 8.70 & 33.33 & 83.30\\

    \hline
     Wafer2Spike~\cite{mishra2024wafer2spike} & 97.17 & 68.49 & 95.43 & 94.32 & 54.67 & 39.37 & 65.21 & 12.63 & 64.29 & 8.24 & 1.09 & 2.08 & 0 & 0 & 89.86 \\
     %hline
    WaferSegClassNet~\cite{nag2022wafersegclassnet} & 97.90 & 72.82 & 99.59 & 99.91 & 73.61 & 0.56 & 59.67 & 0 & 2.20 & 0 & 0 & 0 & 0 & 0 & 90.40 \\
     
     \hline
     \textbf{Ours} & \textbf{98.40} & \textbf{75.24} & \textbf{98.60} & \textbf{99.91} & \textbf{67.41} & \textbf{69.22} & \textbf{83.61} & \textbf{52.28} & \textbf{96.70} & \textbf{50.87} & \textbf{31.52} & \textbf{56.25} & \textbf{8.70} & \textbf{50.00} & \textbf{93.83} \\
%     \hline
%     \textbf{Ours (Yu-Cheng)} & \textbf{99.39} & \textbf{81.52} & \textbf{98.81} & \textbf{99.91} & \textbf{74.33} & \textbf{57.18} & \textbf{76.18} & \textbf{33.68} & \textbf{89.56} & \textbf{8.67} & \textbf{5.43} & \textbf{25.00} & \textbf{0} & \textbf{0} & \textbf{94.44} \\
%     Ours (w/o RC-Attn) & 98.45 & 83.30 & 52.46 & 31.17 & 93.61 \\
%     Ours (Spat. Attn) & 98.54 & 82.66 & 57.38 & {40.26} & 93.59 \\
%     Ours (Chan. Attn) & 98.65 & 83.43 & {61.61} & {35.06} & {93.94} \\
     \hline
     \hline
     \# of Samples & 33,418 & 4,563 & 2,427 & 2,287 & 1,114 & 1,072 & 848 & 285 & 182 & 173 & 92 & 48 & 23 & 6 & 46,538\\
     \hline

    \end{tabular}
    %\captionsetup{font={bf,Large}}   
    \label{tab:UMC_per_defect}
\end{table*}

\begin{table*}[!t]
    \caption{Ablation study for the loss terms for ReCAME-Net on IC-Defect-14 and ImageNet-LT}
    \vspace{-0.1cm}
    \centering
    %\footnotesize
    \footnotesize
    %\begin{tabular}{|llc|c|c|c|c|c|c|}
    \begin{tabular}{|ccccc|c|c|c|c|c|c|c|c|c|}
    %\begin{tabular}{|c|c|c|c|c|c|c|c|c|c|c|c|c|c|}
    \hline 
    \multicolumn{5}{|c|}{Loss terms} & \multicolumn{5}{|c|}{(A) IC-Defect-14 (Acc. \%)} & \multicolumn{4}{c|}{(B) ImageNet-LT (Acc. \%)} \\
    \hline
    $\mathcal{L}_\mathrm{ARB}$ & $\mathcal{L}_\mathrm{HCM}$ &
    $\mathcal{L}_\mathrm{cont}$ &
    $\mathcal{L}_\mathrm{cen}$ &
    $\mathcal{L}_\mathrm{KD}$ & Head & Many & Med. & Few & Whole & Many & Med. & Few & Avg. \\
    & & & & 
    & ($>10^4$)
    & ($10^3$--$10^4$) 
    & ($10^2$--$10^3$) 
    & ($<10^2$)
    &
    & ($>10^2$) 
    & ($20$--$10^2$)
    & ($<20$)
    & \\
    \hline
    - & v & v & v & v & 96.96 & 82.32 & 56.28 & 29.87 & 92.33 & 74.8 & 46.9 & 15.8 & 53.4 \\
    v & - & v & v & v & 96.07 & 79.28 & 71.10 & 28.40 & 90.89 & 70.3 & 56.1 & 39.0 & 59.2 \\
    v & v & - & v & v & 98.84 & 80.23 & 73.99 & 33.72 & 93.22 & 71.8 & 56.0 & 38.6 & 59.7 \\
    v & v & v & - & v & 98.24 & 81.12 & 71.10 & 33.13 & 92.92 & 70.3 & 56.1 & 40.3 & 59.4 \\
    v & v & v & v & - & 98.84 & 79.77 & 69.95 & 32.54 & 92.98 & 70.1 & 54.3 & 37.7 & 58.1 \\
    v & v & v & v & v & \textbf{98.40} & \textbf{83.77} & \textbf{60.38} & \textbf{40.26} & \textbf{93.83} &\textbf{70.2} & \textbf{56.2} & \textbf{39.8} & \textbf{59.4} \\
    \hline
    %\hline
    %\multicolumn{5}{|c|}{\# of Samples} & 33,418 & 12,311 & 732 & 77 & 46,538 & 7,700 & 9,580 & 2,720 & 20,000 \\
    %\hline
    
    \end{tabular}
    %\captionsetup{font={bf,large}}
    
    \label{tab:ablation_losses}
\end{table*}
\begin{table*}[!t]
    \caption{Ablation study for the attention modules of ReCAME-Net on IC-Defect-14 and ImageNet-LT}
    \vspace{-0.1cm}
    \footnotesize
    \centering
    \small
    \begin{tabular}{|cc|c|c|c|c|c|c|c|c|c|c|c|}
    \hline
    \multicolumn{3}{|c|}{Attention Module} & \multicolumn{5}{c|}{(A) IC-Defect-14 (Acc. \%)} & \multicolumn{4}{c|}{(B) ImageNet-LT (Acc. \%)} \\ 
    \hline
    \multicolumn{2}{|c|}{CBAM} & RC-Attn. & Head & Many & Med. & Few & Whole & Many & Med. & Few & Whole \\
    \cline{1-2}
    Channel & Spatial &
    & ($>10^4$)
    & ($10^3$--$10^4$) 
    & ($10^2$--$10^3$)
    & ($<10^2$) &
    & ($>10^2$) 
    & ($20$--$10^2$) 
    & ($<20$)
    &
    \\
    \hline
    - & - & - & 98.45 & 83.30 & 52.46 & 31.17 & 93.61 & 69.9 & 55.5 & 40.10 & 59.0 \\
    v & - & - & 98.65 & 83.43 & 61.61 & 35.06 & 93.94 & 69.8 & 54.3 & 38.4 & 58.1 \\
    - & v & - & {98.54} & 82.66 & 57.38 & {40.26} & 93.59 & {70.1} & {56.0} & {40.0} & {59.2} \\
    v & v & -- & 97.27 & 81.99 & 83.89 & 44.97 & 92.66 & 70.1 & 55.5 & 39.4 & 59.1\\
    
     - & - & v & \textbf{98.40} & {\textbf{83.77}} & {\textbf{60.38}} & {\textbf{40.26}} & {\textbf{93.83}} & \textbf{{70.2}} & \textbf{{56.2}} & \textbf{39.8} & \textbf{{59.4}} \\
    \hline
    \hline
    \multicolumn{3}{|c|}{\# of Samples} & 33,418 & 12,311 & 732 & 77 & 46,538 & 7,700 & 9,580 & 2,720 & 20,000 \\
    \hline
    \end{tabular}
    %\captionsetup{font={bf,large}}
    \label{tab:ablation_attn}
\end{table*}

\subsection{Performance Evaluation}
\label{section: performance}
%\textcolor{blue}
%{
We first evaluate our proposed ReCAME-Net by comparing it with state-of-the-art (SOTA) models designed for highly imbalanced and long-tailed data classification, including NCL~\cite{li2022nested} and MDCS~\cite{Zhao_2023_ICCV}, on the IC-Defect-14 dataset. Following this, we assess ReCAME-Net’s generalization capability by benchmarking it against numerous SOTA models, such as RIDE~\cite{wang2020long}, ACE~\cite{cai2021ace}, and SHIKE~\cite{jin2023long}, on ImageNet-LT and iNaturalist-2018.  All experimental results are reported in Top-1 accuracy and presented in tabular format. To provide a more granular analysis, we follow the evaluation protocol in~\cite{liu2019large} by reporting model performance across four data subsets, categorized based on the number of training samples per class: (i) head classes,  (ii) many-shot classes,  (iii) medium-shot classes, and (iv) few-shot classes. Note that in the following tables, performance values of our default model are highlighted in \textbf{boldface} and results better than ours are indicated in \textit{italic}.
%}

%\subsubsection{IC-Defect-14}
%\label{section: result on UMC}

% v1
% \textbf{UMC dataset}: In this paragraph, we compare the results of ReCAME-Net with NCL~\cite{li2022nested}, and MDCS~\cite{Zhao_2023_ICCV}. Furthermore, we also exhibit the performance of ReCAME-Net combined with various attention modules. The comparison results on UMC dataset are shown in Table~\ref{tab: result on UMC}. From Table~\ref{tab: result on UMC}, the following conclusions can be inferred. First, our approaches outperform existing long-tailed state-of-the-art methods by at least roughly 0.4\% in total accuracy. Second, when ReCAME-Net is equipped with channel attention or RC attention, the results are particularly favorable. Third, even though the overall performance with RC attention is inferior to that with channel attention, we suspect that this may arise because a lot of  defects occupy a tiny area in the entire image, making the improvement of RC attention less pronounced.

% v2

%\textcolor{blue}{
Table~\ref{tab:umc_inat_imagenet} presents a comparative analysis of ReCAME-Net and eight state-of-the-art (SOTA) models for long-tailed and imbalanced data classification, including RIDE~\cite{wang2020long},  TSC~\cite{li2022targeted}, FBL~\cite{li2022feature}, NCL~\cite{li2022nested}, SBCL~\cite{hou2023subclass}, MDCS~\cite{Zhao_2023_ICCV}, SHIKE~\cite{jin2023long}, and BalPoE~\cite{aimar2023balanced}. The evaluation was conducted on the testing set of IC-Defect-14 and two standard open datasets: iNaturalist-2018~\cite{van2018inaturalist} and ImageNet-LT~\cite{liu2019large}. The table reveals the following key observations. 

First, while long-tailed classification SOTAs consistently outperform ResNet and its variants on iNaturalist-2018 and ImageNet-LT, they underperform on the IC-Defect-14 dataset. ResNet and its variants achieve superior performance across nearly all subgroups of IC-Defect-14, suggesting that existing long-tailed classification models, such as RIDE, TSC, FBL, SBCL, and BalPoE, struggle when applied to real-world IC defect image datasets characterized by the multi-cluster head class property. This limitation underscores the difficulty of adapting generic long-tailed learning techniques to industrial defect classification, where intra-class variations in the head class are often significant.

Second, ReCAME-Net outperforms previous SOTAs on IC-Defect-14 while maintaining competitive performance on general long-tailed datasets, iNaturalist-2018 and ImageNet-LT, demonstrating its strong generalization capability in imbalanced classification tasks. This result reflects that while ReCAME-Net was originally designed for real-world highly imbalanced datasets without prior knowledge of data distribution, it still performs comparably against methods that explicitly leverage prior distributional information. This robustness and stability indicate that ReCAME-Net effectively balances adaptability to real-world data distributions with competitive accuracy on standard long-tailed datasets. 

Third, compared to the baseline performance achieved by ResNet, ReCAME-Net improves classification accuracy across many-shot, medium-shot, and few-shot categories, with only a slight decrease in accuracy for the head class subset.\footnote{Among the four subgroups shown in Table~\ref{tab:umc_inat_imagenet}, the head class subset consists exclusively of normal regular images (Class-0), whereas the remaining three subsets contain defect images.} In contrast, long-tailed classification SOTAs such as RIDE, FBL, and SHIKE exhibit notable performance degradation in medium-shot and few-shot classes when applied to IC-Defect-14, highlighting their limitations in handling highly imbalanced real-world classification tasks. This observation further emphasizes the inherent trade-off in most long-tailed classification approaches: models that optimize for head class performance tend to struggle with tail class accuracy. While prior SOTAs have shown improvements in head class accuracy and many-shot subsets, they still face challenges in achieving comparable gains for minority classes, particularly the few-shot subset.

Overall, the results in Table~\ref{tab:umc_inat_imagenet} validate the effectiveness of ReCAME-Net for highly imbalanced IC defect image classification, making it particularly well-suited for real-world high-yield manufacturing applications where accurate defect detection across different class distributions is critical.
%}

\iffalse
\textcolor{red}{
ReCAME-Net demonstrates slightly lower average accuracy compared to MDCS across the entire dataset, it notably outperforms both MDCS and NCL on the medium-shot and few-shot subsets. This observation underscores the trade-off between performance improvement in head classes and that in tail classes for existing methods, thereby highlighting the weaknesses of previous approaches.  %This initial observation implies two points.  
Second, even latest state-of-the-art methods like MDCS and NLC still have not yet discovered and thus regarded the data properties found in this real-world IC manufacturing data. Third, consequently, while previous SOTAs show improvements in the classification accuracy of the head class subset and the many-shot subset, it is still challenging for them to achieve similar gains in minority classes, especially the few-shot subset. Lastly, despite the stable and robust performance of all ReCAME-Net variants, we recommend equipping ReCAME-Net with RC-Attn as the default setting, as it achieves optimal performance on defect image subsets\footnote{Among the four subsets shown in Table \ref{tab:result_on_UMC}, the subset of head class contains only normal regular images of Class-0, while the other three subsets are comprised of defect images.}. }
\fi

%\textcolor{blue}{
Table~\ref{tab:UMC_per_defect} presents a per-defect accuracy comparison of various recent long-tailed classification SOTAs and our proposed method. This table also highlights the following findings.
First, RIDE~\cite{wang2020long} introduced a parallel multi-expert classifier framework, which later inspired techniques such as NCL~\cite{li2022nested}, MDCS~\cite{Zhao_2023_ICCV}, and BalPoE~\cite{aimar2023balanced}. While these methods integrate multiple expert classifiers, they fail to account for real-world data complexities found in datasets like IC-Defect-14. Specifically, RIDE struggles significantly with tail categories, while MDCS improves accuracy on medium-shot and few-shot defects but at the cost of drastically reduced accuracy for Class-0 (non-defective samples). As shown in Table~\ref{tab:UMC_per_defect}, this category of multi-expert approaches fails to handle both minority defect classes (e.g., defect codes 132, 14, 111, 153, 129, 113, 158, and 37) and major defect categories (e.g., defect codes 114, 116, and 126), making them suboptimal for industrial defect classification.

Second, TSC~\cite{li2022targeted} applies contrastive learning to increase the logit score separation between classes to improve classification accuracy. However, its performance on IC-Defect-14 remains ineffective, as it does not sufficiently address the unique challenges posed by the dataset. In contrast, ReCAME-Net employs contrastive learning directly at the feature vector level, effectively pulling feature points of the same class closer while pushing apart those of different classes. This design directly mitigates the challenges of IC-Defect-14, such as multi-cluster head classes, large intra-class diversity, and high inter-class similarity, making it more effective in real-world settings.
Third, SBCL~\cite{hou2023subclass}  builds upon TSC by attempting to increase inter-class spacing in the feature space. However, its performance is even worse than TSC’s. While SBCL applies a $k$-means clustering strategy to compute class centers and adjust inter-class distances, it lacks a mechanism to condense intra-class feature distributions. As a result, it performs the worst among all evaluated methods on IC-Defect-14, struggling to handle both head and tail classes.
Fourth, FBL adjusts classification boundaries by correcting logit scores using the formula $z^b_j = z_j - \alpha \frac{\lambda_j}{|\mathbf{f}|}$. However, FBL does not incorporate adaptation mechanisms tailored to real-world IC defect image distributions, making it ineffective, particularly for tail classes. Although FBL performs well on standard long-tailed datasets, its inability to model the structural complexities of IC defects results in suboptimal classification performance on IC-Defect-14.
Fifth, while ReCAME-Net achieves strong performance on IC-Defect-14, two recent classification methods for IC-industry, e.g. Wafer2Spike~\cite{mishra2024wafer2spike} and WaferSegClassNet~\cite{nag2022wafersegclassnet}, are not sufficiently robust to handle IC-Defect-14.

These findings collectively indicate that while existing long-tailed classification methods perform well on open datasets like ImageNet-LT and iNaturalist-2018, they are insufficient for real-world IC defect classification tasks. This underscores the necessity of models like ReCAME-Net, which are specifically designed to handle the unique challenges of industrial defect datasets. NCL~\cite{li2022nested} employs a multi-classifier framework that incorporates majority voting, knowledge alignment, and a relearning strategy for difficult categories, making it one of the better-performing models for real-world datasets. The proposed ReCAME-Net adopts a similar concept but enhances it with attention-based feature refinement and ARB loss to mitigate minority class collapse, leading to superior classification accuracy across all defect categories. The results in Table~\ref{tab:UMC_per_defect} clearly demonstrate ReCAME-Net’s effectiveness in learning and distinguishing defect categories in a highly imbalanced industrial setting, making it well-suited for real-world high-yield manufacturing applications. 
%}

\iffalse
In this paragraph, we compare the results of ReCAME-Net with NCL~\cite{li2022nested}, and MDCS~\cite{Zhao_2023_ICCV}. Furthermore, we also exhibit the performance of ReCAME-Net combined with various attention modules. The comparison results on UMC dataset are shown in Table~\ref{tab: result on UMC}. From Table~\ref{tab: result on UMC}, the following conclusions can be inferred. First, our method outperforms NCL~\cite{li2022nested} but falls behind MDCS~\cite{Zhao_2023_ICCV} by approximately 1\% in total accuracy. However, the superior performance of MDCS stems from a significant bias towards majority classes. Second, even though the overall performance with RC attention is inferior to that with channel attention, we analyze that this may be attributed to the fact that many defects occupy a small area in the entire image, thereby making the improvement of RC attention less pronounced. 
\fi

\begin{figure}
    \centering
    \includegraphics[width=0.49\textwidth]{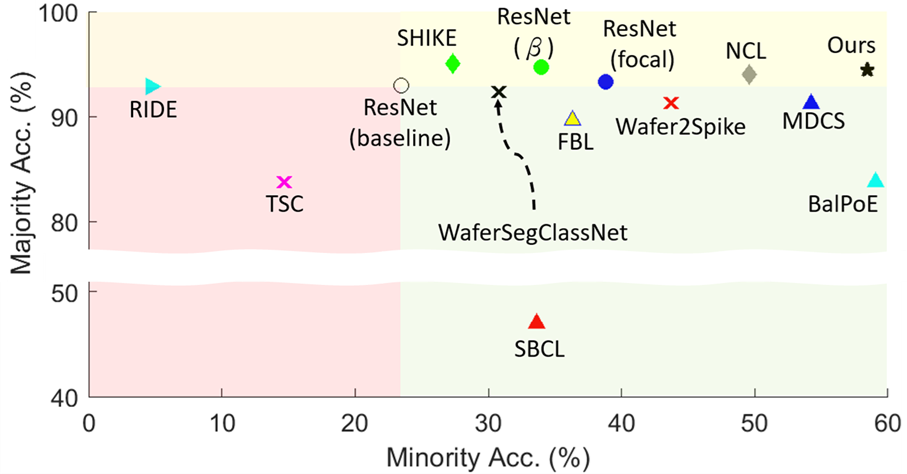}
    \caption{Performance of representative long-tailed classification methods and IC image classification models on the \textbf{IC-Defect-14} dataset, comparing majority and minority class accuracy against the baseline ResNet model (marked by the circle. The figure shows that only a few methods appear in the top-right quadrant, indicating that most prior SOTAs fail to effectively improve defect category classification. While some approaches enhance minority class performance at the cost of majority class accuracy, others prove inadequate for IC-Defect-14 classification. In contrast, our proposed ReCAME-Net achieves improvements in both majority and minority classes simultaneously.}
    \label{fig:tradeoff_UMC}
\end{figure}
\begin{figure*}
    \centering
    \includegraphics[width=0.90\textwidth]{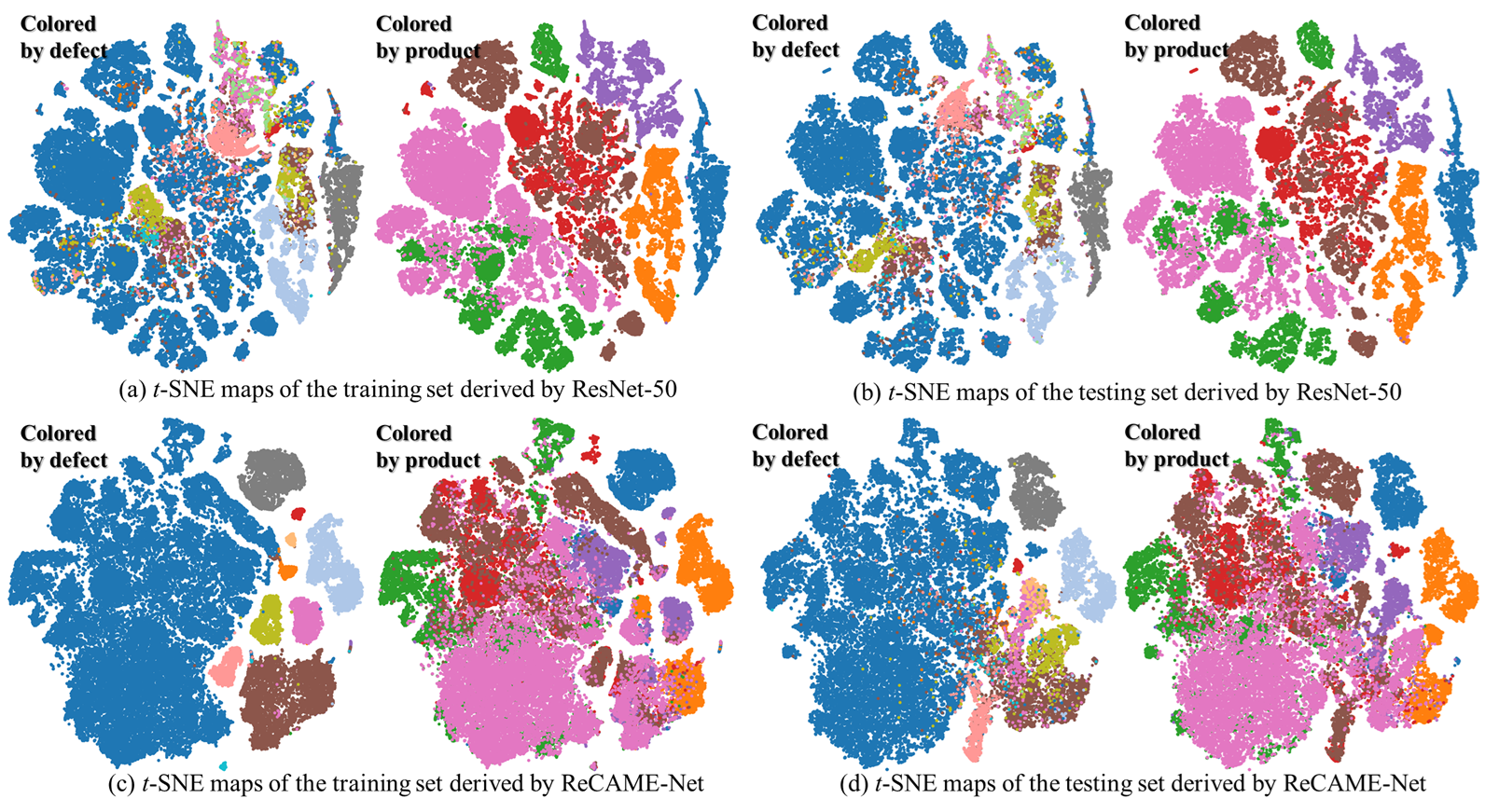}%{figures/ICME_our_tsne2.png}
    \caption{\textit{t}-SNE visualizations of IC-Defect-14 feature distributions generated by ResNet-50 and the proposed ReCAME-Net. Each subfigure contains a pair of \textit{t}-SNE maps: one color-coded by defect category and the other by production line.
\textbf{(a)} and \textbf{(b)}: ResNet-50’s feature space is influenced by production line attributes, suggesting that traditional methods struggle to suppress production line biases, ultimately compromising defect detection performance.
\textbf{(c)} and \textbf{(d)}: In contrast, ReCAME-Net effectively clusters features by defect category while increasing the separation between different defect classes. This allows ReCAME-Net to prioritize defect classification over production line variations, contributing to its state-of-the-art performance on IC-Defect-14.}
    \label{fig:ourtsne}
\end{figure*}

%\textcolor{blue}{
Following the representation proposed in~\cite{cai2021ace}, we visualize in Fig.~\ref{fig:tradeoff_UMC} the performance trade-off between majority and minority class classification accuracy for each model on IC-Defect-14. This figure illustrates the challenge of maintaining a balance between these two performance aspects while ensuring strong classification accuracy across both frequent and rare defect categories. ReCAME-Net demonstrates greater robustness against this trade-off than the other SOTA models, emphasizing its effectiveness in real-world highly imbalanced IC defect image classification tasks. Unlike prior methods that often favor majority classes at the cost of minority class performance, ReCAME-Net maintains a stable balance between the two, ensuring improved defect recognition across all categories. 
In Fig.~\ref{fig:tradeoff_UMC}, the majority class accuracy is defined as the proportion of correctly predicted samples in the head and many-shot classes, while the minority accuracy is similarly determined based on the medium-shot and few-shot classes, as outlined in~\cite{cai2021ace}:
\begin{eqnarray}
    \mathrm{Acc_{major}} &=& \frac{\mathrm{Correct}_\mathrm{Head} + \mathrm{Correct}_\mathrm{Many}}{\mathrm{Total}_\mathrm{Head} + \mathrm{Total}_\mathrm{Many}}\mbox{,}\\
    \mathrm{Acc_{minor}} &=& \frac{\mathrm{Correct}_\mathrm{Medium} + \mathrm{Correct}_\mathrm{Few}}{\mathrm{Total}_\mathrm{Medium} + \mathrm{Total}_\mathrm{Few}}
    \mbox{.}
    \label{eq:perftradeoff}
\end{eqnarray} 

%\textcolor{red}{Additionally, Fig.~\ref{fig:ourtsne} provides \textit{t}-SNE visualizations of IC-Defect-14 feature distributions as generated by ResNet and ReCAME-Net. The visualization clearly shows that ReCAME-Net effectively condenses intra-class clusters (highlighted in blue) while simultaneously enlarging the margins between different defect categories (highlighted in red). This superior feature separation contributes significantly to ReCAME-Net’s state-of-the-art performance on real-world IC defect classification tasks, ensuring better discrimination of highly imbalanced defect types. }
%\textcolor{blue}{
Fig.~\ref{fig:ourtsne} presents \textit{t}-SNE visualizations of feature distributions in IC-Defect-14, comparing ResNet-50 and ReCAME-Net. Each subfigure consists of two images: the left image is color-coded by defect type, while the right image is color-coded by product line ID. Figs.~\ref{fig:ourtsne}(a) and \ref{fig:ourtsne}(b) depict the \textit{t}-SNE maps of the training and testing subsets derived from ResNet-50, while Figs.~\ref{fig:ourtsne}(c) and \ref{fig:ourtsne}(d) show the corresponding maps generated by ReCAME-Net.

These visualizations reveal several key insights. First, in Figs.~\ref{fig:ourtsne}(a) and \ref{fig:ourtsne}(b), the right images indicate that features extracted by ResNet-50 tend to cluster by product line rather than defect type. This suggests that despite being trained for defect classification, ResNet-50 remains influenced by production line characteristics, potentially compromising its robustness and leading to reduced defect detection performance, as demonstrated in Table~\ref{tab:UMC_per_defect}.
In contrast, as shown in the left-images of Figs.~\ref{fig:ourtsne}(c) and \ref{fig:ourtsne}(d), ReCAME-Net effectively clusters features based on defect categories while reducing the influence of product line attributes. This results in a feature space that prioritizes defect classification rather than production line biases, underscoring ReCAME-Net’s ability to enhance defect detection accuracy.
%}

%\textcolor{blue}{
Finally, the visualizations clearly illustrate ReCAME-Net’s ability to compactly cluster intra-class features (highlighted in blue) while enlarging the separation margins between distinct defect categories (highlighted in red). This enhanced feature separation is a key factor in ReCAME-Net’s state-of-the-art performance in real-world IC defect classification. By effectively distinguishing highly imbalanced defect types, ReCAME-Net significantly improves classification accuracy, ensuring greater robustness and reliability in practical defect detection applications.

\subsection{Ablation Study}

%\textcolor{blue}
{
Our ablation study consists of two sets of experiments. The first, shown in Table~\ref{tab:ablation_losses}, validates the contribution of each loss term, while the second, shown in Table~\ref{tab:ablation_attn}, evaluates the effectiveness of different attention modules. In cases where $\mathcal{L}_\mathrm{ARB}$ is removed, we substitute it with a traditional cross-entropy loss.
For a comprehensive evaluation, we conducted our loss-based ablation study on both IC-Defect-14 and ImageNet-LT. While IC-Defect-14 exhibits real-world multi-cluster defect distributions, ImageNet-LT is a widely used manually curated natural image dataset where each class contains only a single data cluster. The results in Table~\ref{tab:ablation_losses} highlight the following key observations:
First, ARB-loss $\mathcal{L}_\mathrm{ARB}$ effectively prevents minority collapse and improves classification accuracy on the medium-shot and few-shot classes by at least 10\%. Without ARB-loss, although the model increases classification accuracy for majority classes to 74.8\%, it significantly reduces accuracy for minority classes.
Second, HCM-loss $\mathcal{L}_\mathrm{HCM}$ improves classification accuracy for hard categories, contributing approximately 0.2\% to overall performance.
Third, KD-loss $\mathcal{L}_\mathrm{KD}$ aligns predictions across multiple experts, leading to a more robust consensus in classification results.
Fourth, while the two metric learning losses, contrastive loss ($\mathcal{L}_\mathrm{cont}$) and center loss ($\mathcal{L}_\mathrm{cen}$), show limited impact on ImageNet-LT due to its single-cluster category structure, they still positively contribute to classification performance, particularly in datasets like IC-Defect-14, where classes exhibit large intra-class variance and multiple cluster formations.
}

{Table~\ref{tab:ablation_attn} presents the ablation study evaluating the impact of different attention modules on ReCAME-Net's performance. The results indicate that RC-Attn achieves the best overall performance across both IC-Defect-14 and ImageNet-LT. Although the channel attention module yields slightly better results on IC-Defect-14, its performance drop on ImageNet-LT reveals its weakness in generalizability. These findings highlight the effectiveness of RC-Attn in enhancing model robustness. Based on these results, as stated in Sec.~\ref{section: performance}, we recommend RC-Attn as the default attention module for ReCAME-Net to ensure optimal performance on highly imbalanced and long-tailed classification tasks.

%\newpage

\subsection{Analysis of Computational Costs and Backbone Selection}

Table~\ref{tab:computecost} lists the computational complexity, training time, and resource requirements for the 2-expert ReCAME-Net and its variants. Due to the use of a multi-expert design combined with the more complex RC-Attention module, ReCAME-Net consumes higher parameter counts and computational complexity than a conventional ResNet-50. Although ReCAME-Net achieves strong performance at a considerable computational cost compared with ResNet-50, its per-image inference time of 71.95 ms on an NVIDIA RTX 3050 GPU corresponds to approximately 7--12 ms on an RTX 4090 and 8--14 ms on an A100 GPU, and thus remains reasonably fast and acceptable.

As discussed above, Since ReCAME-Net adopts a multi-expert framework, it inherits the common limitations of such classifiers, including longer training times and higher memory costs, as shown in Table~\ref{tab:computecost}. Consequently, for AOI-oriented classification, an important direction for future research is to reduce the computational and memory demands of defect classification algorithms while maintaining high performance on highly imbalanced real-world data. Achieving this would enable models like ReCAME-Net to be more broadly applicable to IC defect classification for output quality assessment.

%====以下是Table VII的完整表格，先註解掉
\iffalse
\begin{table*}[!t]
\color{blue}
    \caption{Comparison of computational cost and model complexity of the 2-expert ReCAME-Net and its variants. Note that the per-image inference time was measured on an NVIDIA 3050 GPU}
    \centering
    \scriptsize
    \begin{tabular}{|c|c||c|c|c|c|c|}
    \hline
       \multicolumn{2}{|c||}{Model Architecture (2-experts) } & Training Time  & Inference time & Inference Time & \# Params  & FLOPS \\
       \cline{1-2}
       Backbone & Attention & (per-epoch) & (per-image) & (normalized) & (M) & (G) \\
       \hline
       \hline
       ResNet-50  & -- &  36m 03s  & 23.36 ms & 2.07x & 47.02 & 8.26 \\ 
       ResNet-50  & Channel & 49m 19s  & 34.49 ms & 3.05x & 52.08 & 8.27 \\
       ResNet-50  & Spatial & 42m 22s  & 30.22 ms & 2.68x & 47.02 & 8.27 \\
       ResNet-50  & CBAM & 55m 54s  & 41.77 ms & 3.70x & 52.08 & 8.28 \\
       ResNet-50  & RC-Attn. &  70m 22s & 71.95 ms & 6.37x & 52.08 & 8.30 \\
       \hline
       \hline
       ResNet-34  & RC-Attn. &  42m 47s & 65.21 ms & 5.78x & 42.89 & 7.36 \\
       ResNet-50  & RC-Attn. & 70m 22s  & 71.95 ms & 6.37x & 52.08 & 8.30 \\
       ResNet-101  & RC-Attn. & 123m 32s  & 145.09 ms & 12.85x & 94.56 & 15.80 \\
       \hline
       %\multicolumn{2}{|c|}{Baseline (original ResNet-50)}  &   & -- & -- & -- & -- 
       %\hline
       \hline
       \multicolumn{2}{|c||}{ResNet-50 (Baseline)}  &  20m 36s & 11.29 ms & 1x & 23.51 & 4.13 \\
       \hline        
    \end{tabular}
    \label{tab:computecost}
\end{table*}
\fi
%====以上是Table VII的完整表格，先註解掉
\begin{table}[!t]
    \caption{Comparison of computational cost and model complexity of the 2-expert ReCAME-Net and its variants. Note that the per-image inference time was measured on an NVIDIA 3050 GPU}
    \centering
    \scriptsize
    \begin{tabular}{|c|c||c|c|c|c|}
    \hline
       \multicolumn{2}{|c||}{Architecture (2-experts) } & Training  & Inference & \# Params  & FLOPS \\
       \cline{1-2}
       Backbone & Attention & (per-epoch) & (per-image) & (M) & (G) \\
       \hline
       \hline
       ResNet-50  & -- &  36m 03s  & 23.36 ms & 47.02 & 8.26 \\ 
       ResNet-50  & Channel & 49m 19s  & 34.49 ms & 52.08 & 8.27 \\
       ResNet-50  & Spatial & 42m 22s  & 30.22 ms & 47.02 & 8.27 \\
       ResNet-50  & CBAM & 55m 54s  & 41.77 ms & 52.08 & 8.28 \\
       ResNet-50  & RC-Attn. &  70m 22s & 71.95 ms & 52.08 & 8.30 \\
       \hline
       \hline
       ResNet-34  & RC-Attn. &  42m 47s & 65.21 ms & 42.89 & 7.36 \\
       ResNet-50  & RC-Attn. & 70m 22s  & 71.95 ms & 52.08 & 8.30 \\
       ResNet-101  & RC-Attn. & 123m 32s  & 145.09 ms & 94.56 & 15.80 \\
       \hline
       %\multicolumn{2}{|c|}{Baseline (original ResNet-50)}  &   & -- & -- & -- & -- 
       %\hline
       \hline
       \multicolumn{2}{|c||}{ResNet-50 (Baseline)}  &  20m 36s & 11.29 ms & 23.51 & 4.13 \\
       \hline        
    \end{tabular}
    \label{tab:computecost}
\end{table}

\begin{table}[!t]
    \caption{Comparison of ReCAME-Net performance on IC-Defect-14 using different feature extraction backbones}
    \centering
    %\Large
    \scriptsize
    \begin{tabular}{|l||c|c|c|c|c|}
    \hline 
     & \multicolumn{5}{c|}{IC-Defect-14} \\ 
     \cline{2-6}
     Backbone & Head & Many & Med. & Few & Avg. \\
      & \scriptsize{($>10^4$)} & 
     \scriptsize{($10^3$--$10^4$)} &
     ($10^2$--$10^3$) & ($<10^2$) &  \\
     \hline
     ResNet-50 & 98.40 & 83.77 & 60.38 & 40.26 & 93.83 \\
     \hline
     ResNet-34 & 98.47 & 83.70 & 73.01 & 29.58 & 93.76 \\
     \hline
     ResNet-101 & 97.99 & 82.02 & 79.96 & 35.50 & 93.25 \\
     \hline
     \hline
     \# of Samples & 33,418 & 12,311 & 732 & 77 & 46,54 \\
     \hline        
     
    \end{tabular}
    %\captionsetup{font={bf,Large}}   
    \label{tab:bkboneperformance}
\end{table}

\subsection{Classification performance on open dataset}

Table~\ref{tab:wm811k} shows ReCAME-Net's performance on the widely used WM-811K~\cite{wm811k} dataset. In this experiment, ReCAME-Net employed the same training loss weights determined using the validation set of IC-Defect-14. Consequently, the results, presented in Table~\ref{tab:wm811k} and Tables~\ref{tab:umc_inat_imagenet} and~\ref{tab:UMC_per_defect}, demonstrate the superiority and generalizability of ReCAME-Net across both the IC-Defect-14 and WM-811K datasets. Finally, Table~\ref{tab:wm811k} shows that previous methods, compared with ResNet-50, may tend to sacrifice the accuracy of the majority class (\textit{i.e.}, ``None-defect'') to improve the accuracy of minority classes, whereas ReCAME-Net improves the accuracy of almost all classes.

\begin{table*}[]
    \caption{Classification accuracy (\%) of different methods on WM-811k. (A) Machine-learning based methods. (B) Previously proposed deep-learning based methods. (C) ResNet-50 and our ReCAME-Net evaluated on the same training/validation/test sets. Note that the performance values in (A) and (B) are those reported in the original source papers}
    \centering
    \scriptsize
    \begin{tabular}{|c|c|c||c|c|c|c|c|c|c|c|c|c|}
        \hline
        & Methods & Data Splitting & \multicolumn{9}{c|}{Defect Types} & Whole\\
        \cline{4-12} 
        
         & & (Train:Val.:Test) & None & Edge-Ring & Edge-Local & Center & Local & Scratch & Random & Donut & Near-full & Dataset\\
        \hline

        %WaferSegClassNet & nn & er & el & cc & ll & ss & rr & dd & nf & avg \\ \hline
        & DTE-WMFPR~\cite{piao2018decision} & N.A. & 100 & 69.9 & 58.6 & 92.5 & 62.1 & 68.6 & 91.0 & 83.0 & N.A. & 78.5 \\
        (A) & WMDPI-SVE~\cite{saqlain2019voting} & 5-fold cross-val. & 100 & 95 & 77 & 86 & 60 & 34 & 94 & 77 & 97 & 95.8 \\ 
        & WMFPR~\cite{wu2014wafer} & 6.7~:~93.3 & 95.7 & 79.7 & 85.1 & 84.9 & 68.5 & 82.4 & 79.8 & 74.0 & 97.9 & 94.6 \\
        \hline
        %\textcolor{red}{XXXXX} & 100 & 97.6 & 90.9 & 95.5 & 88.5 & 88.9 & 95.4 & 93.5 & 91.9 & 93.6 \\ \hline
        
        & T-DenseNet~\cite{shen2019wafer} & N.A. & 85.5 & 66.8 & 81.5 & 64.5 & 100 & 72.6 & 65.5 & 91.2 & 99.3 & 87.7 \\ 
        
        & Wafer2Spike~\cite{mishra2024wafer2spike} & 6~:~1~:~3 & 99 & 96 & 87 & 97 & 84 & 49 & 97 & 97 & 86 & 97 \\
        
        (B) & WaferDSL~\cite{alawieh2020wafer} &         7~:~1~:~2 & 97.77 & 93.40 & 69.93 & 95.68 & 64.40 & 28.73 & 77.78 & 40.00 & 72.50 & 93.84 \\ 
        
        & WaferHSL \cite{wang2022waferhsl}& 6~:~4~:~0 & 99.34 & 98.30 & 85.84 & 96.10 & 78.37 & 84.10 & 90.20 & 89.64 & 98.33 & 98.12 \\ 
        
        & WM-PeleeNet \cite{yu2023wafer} & 6~:~1.5~:~2.5 & 100 & 97.6 & 90.9 & 95.5 & 88.5 & 88.9 & 95.4 & 93.5 & 91.9 & 93.6 \\ 
        %\textcolor{red}{XXXXX} & nn & er & el & cc & ll & ss & rr & dd & nf & avg \\ \hline
        
        \hline \hline 
        & ResNet-50 & 6~:~1~:~3 & 99.88 & 0 & 66.02 & 0 & 66.33 & 0 & 0 & 0 & 0 & 88.50 \\ 
%        & ReCAME-Net & 6~:~1~:~3 & 99.93 & 96.52 & 66.67 & 88.21 & 62.15 & 35.20 & 70.27 & 64.46 & 44.44 & 96.91 \\ \cline{2-13}
        & ReCAME-Net & 6~:~1~:~3 & 99.92 & 96.38 & 64.48 & 89.68 & 68.27 & 37.99 & 72.59 & 70.48 & 62.22 & 97.05 \\ \cline{2-13}
        (C) & \multicolumn{2}{l||}{\# Training samples (60\%)} & 88459 & 5808 & 3113 & 2576 & 2156 & 716 & 520 & 333 & 89 & 103770 \\
        & \multicolumn{2}{l||}{\# Validation samples (10\%)}& 14743 & 968 & 519 & 429 & 359 & 119 & 87 & 56 & 15 & 17295 \\
        & \multicolumn{2}{l||}{\# Test samples (30\%)}& 44229 & 2904 & 1557 & 1289 & 1078 & 358 & 259 & 166 & 45 & 51885 \\
        \hline
    \end{tabular}
    \label{tab:wm811k}
\end{table*}

%% file: TCAD_sec05_conclusion.tex
%\textcolor{blue}
%{
In this paper, we revealed the previously unexplored data properties of a real-world highly imbalanced dataset, IC-Defect-14, involving ``many in few'' and ``few in many'' characteristics, \textit{i.e.}, multi-cluster head classes, large intra-class diversity, and high inter-class similarity. These characteristics pose challenges for existing highly-imbalanced classification techniques when applied to real-world classification tasks in the IC industry. To address these challenges, we proposed ReCAME-Net, a classification network that integrates hard category mining, knowledge distillation design, regional channel attention, and metric learning losses within a multi-expert classification framework. ReCAME-Net effectively ensures a sufficient margin between different defect categories while centralizing intra-class clusters in feature space. Extensive experiments demonstrate that ReCAME-Net achieves optimal performance on the IC-Defect-14 dataset and competes favorably with recent state-of-the-art classification models on two widely-used open datasets. This study contributes to bridging the gap between fields such as the IC fabrication, computer vision, and data science.
%}

%In this paper, we reported the unexplored data properties from a real-world highly imbalanced dataset (UMC dataset), such as multi-cluster head classes and high inter-class similarity, etc. In order to alleviate the impact of these characteristics, we have proposed the ReCAME-Net, which ensures a sufficient margin between inter-class while centralizing intra-class in feature space. Furthermore, ReCAME-Net enhances the quality of Convolutional Neural Network (CNN) features through the Regional Channel Attention (RC-Attn) module. Extensive experiments show that on real-world highly imbalanced dataset: UMC dataset, our ReCAME-Net is superior to existing long-tailed methods (less trade-off). Moreover, it exhibits competitive results on traditional long-tailed open datasets as well. Finally, through this paper, we have bridged the gap between computer vision and other fields like semiconductors and biology, making computer vision more widely used. 

%% file: bio.tex
\vspace{-0.4in}
\begin{IEEEbiography}
[{\includegraphics[width=1in,height=1.25in,clip,keepaspectratio]{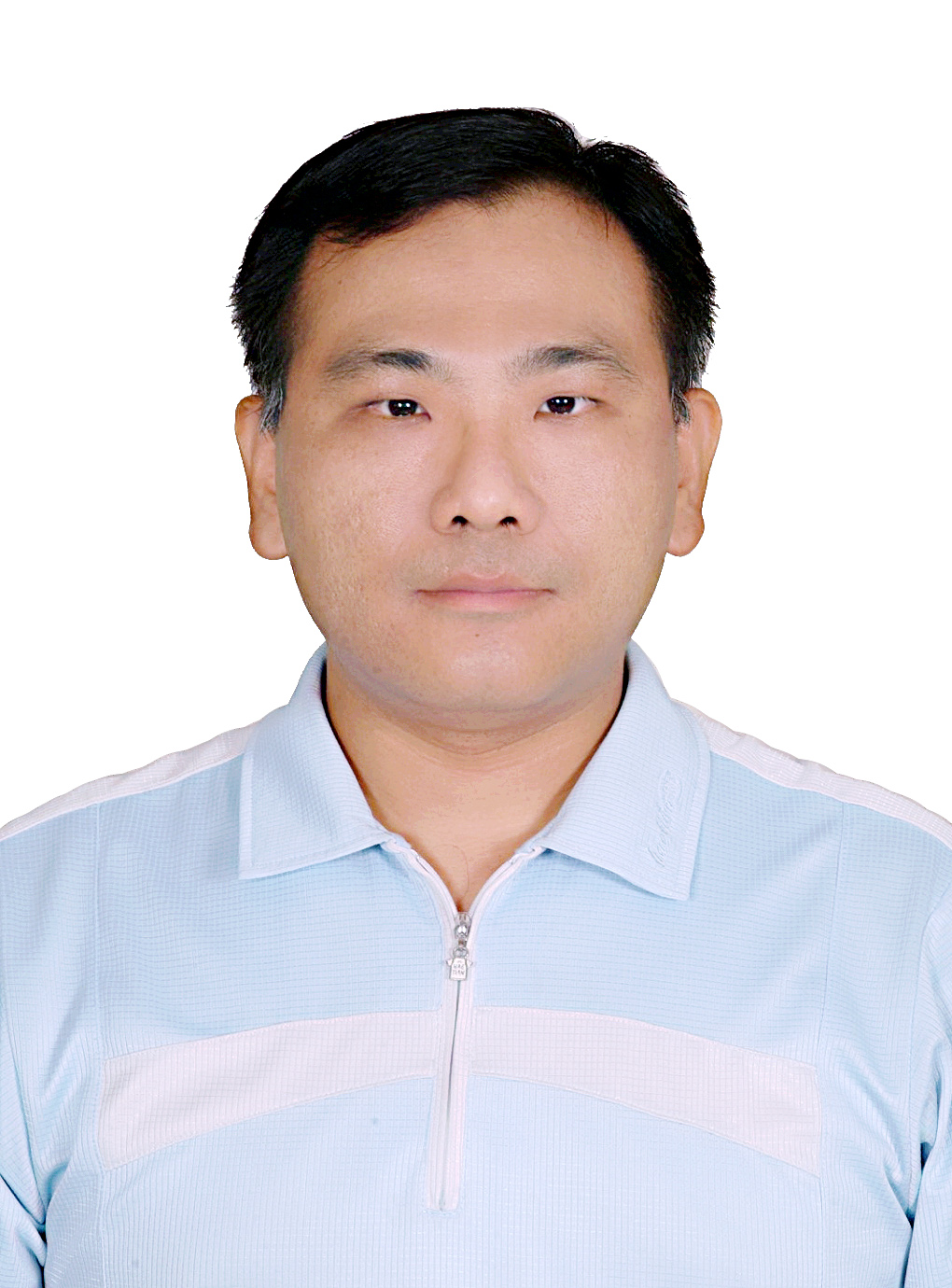}}]
{Hao-Chiang~Shao}
(Senior Member, IEEE) received his Ph.D. degree in electrical engineering from National Tsing Hua University (NTHU), Taiwan, in 2012. 
He is currently an Associate Professor with the Inst. Data Science and Information Computing, NCHU, Taiwan. From 2018 to 2022, he was an Assistant Professor with the Department of Statistics and Information Science, Fu Jen Catholic University, Taiwan. From 2012 to 2017, he was a postdoctoral researcher with the Institute of Information Science, Academia Sinica, involved in \textit{Drosophila} brain research projects; in 2017--2018, he was an R\&D engineer with the Computational Intelligence Technology Center, Industrial Technology Research Institute, Taiwan, taking charges of DNN-based automated optical inspection projects. His research interests include 2D+Z image atlasing, forgery and anomaly detection, big image data analysis, classification methods, and machine learning.
\end{IEEEbiography}

%\newpage

\vspace{-0.4in}
\begin{IEEEbiography}
	[{\includegraphics[width=1in,height=1.25in,clip,keepaspectratio] {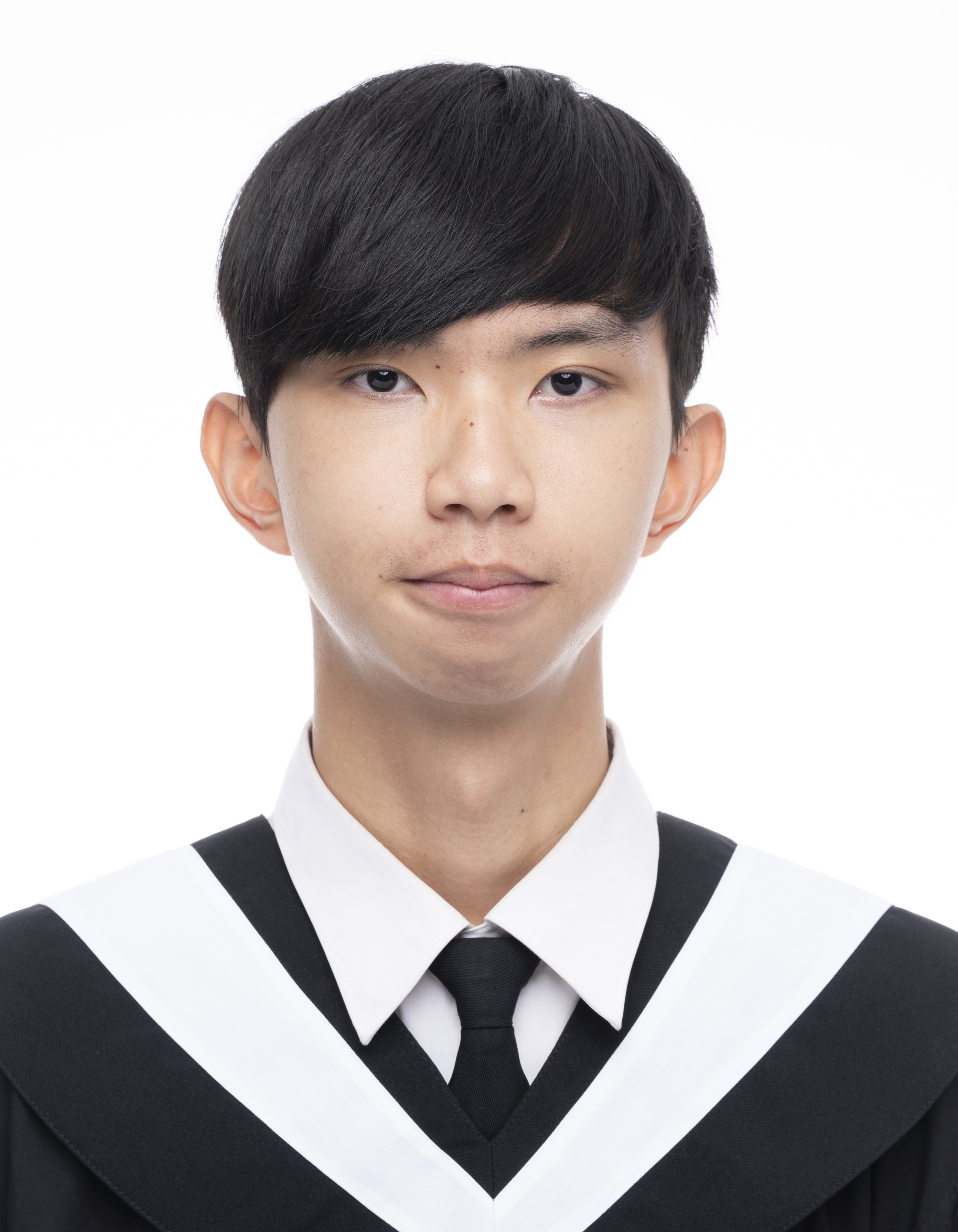}}]
	{Chun-Hao Chang}
	 received his B.S. degree in Electrical Engineering from the National Taipei University of Technology in 2021 and M.S. degree in Electrical Engineering from National Tsing Hua University in January 2024.  
	He joined NovaTek, Hsinchu, Taiwan, as a software engineer in 2024.   His research interests lie in computer vision, machine learning, and visual analytics for IC design for manufacturability.
\end{IEEEbiography}

\vspace{-0.4in}
\begin{IEEEbiography}
	[{\includegraphics[width=1in,height=1.25in,clip,keepaspectratio] {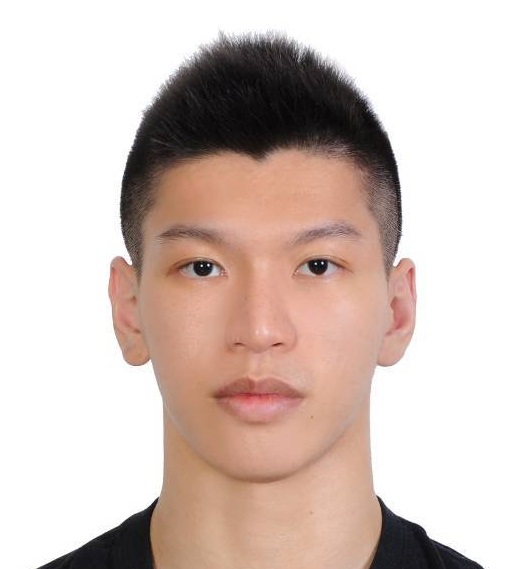}}]
	{Yu-Hsien Lin} received the B.S. degree in computer science from the Fu Jen Catholic University, New Taipei City, Taiwan, in 2020, Taiwan, the M.A. degree in applied statistics from Fu Jen Catholic University, in 2022. He is currently pursuing his Ph.D. degree at the Department of Electrical Engineering of National Tsing Hua University, Hsinchu, Taiwan.
His research interests mainly lie in machine learning, electronic design automation, and computer vision. 
\end{IEEEbiography}

\vspace{-0.4in}
\begin{IEEEbiography}
	[{\includegraphics[width=1in,height=1.25in,clip,keepaspectratio] {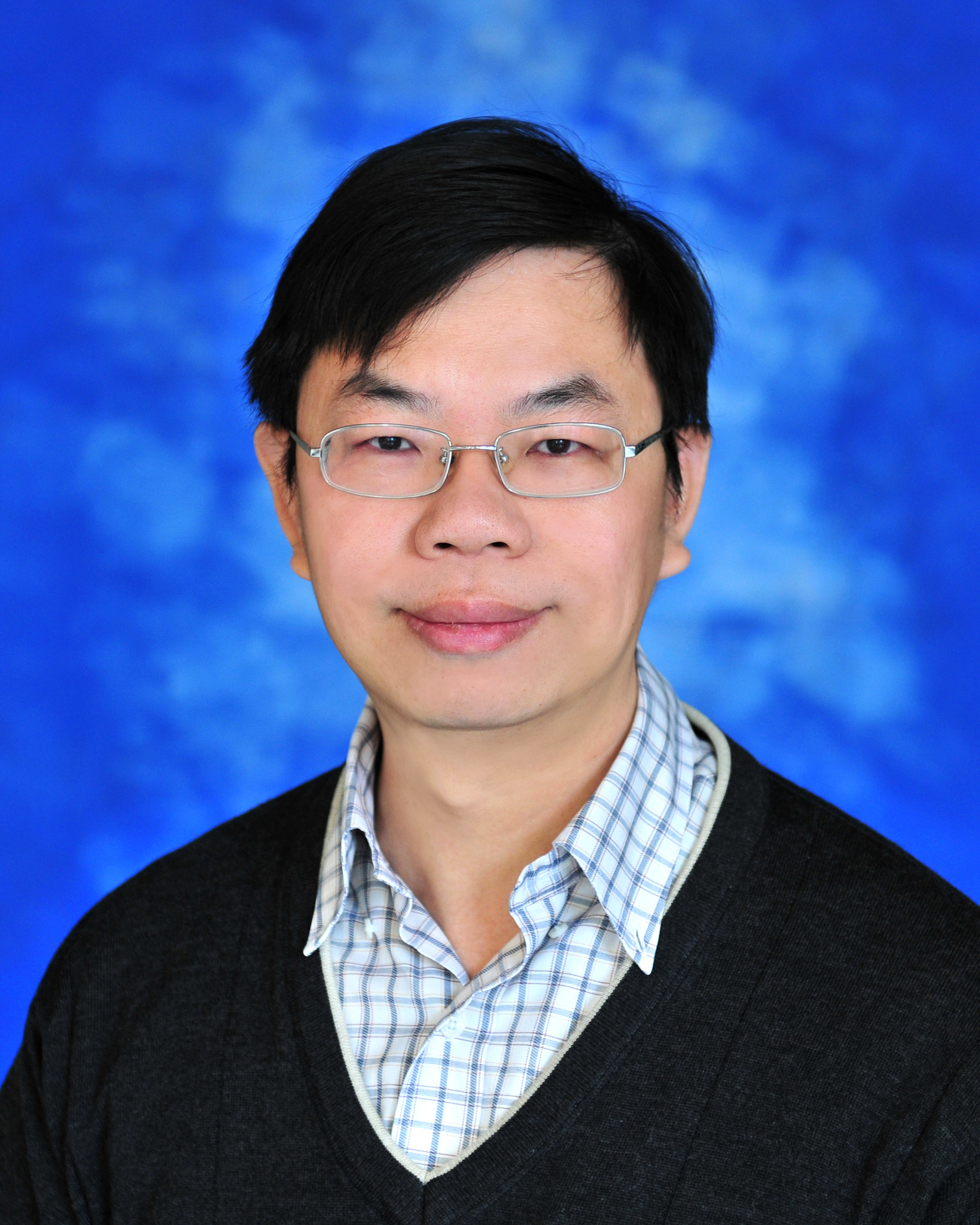}}]
	{Chia-Wen Lin}
	(Fellow, IEEE) received his Ph.D. degree from National Tsing Hua University (NTHU), Hsinchu, Taiwan, in 2000.  
	Dr. Lin is currently a Distinguished Professor with the Department of Electrical Engineering and the Institute of Communications Engineering, NTHU. His research interests include visual signal processing and computer vision.  He served as a Fellow Evaluation Committee member (2021--2023), BoG Members-at-Large (2022--2024), and Distinguished Lecturer (2018--2019) of the IEEE Circuits and Systems Society.   He was Chair of IEEE ICME Steering Committee (2020--2021). He served as TPC Co-Chair of PCS 2022, IEEE ICIP 2019 and IEEE ICME 2010, and General Co-Chair of PCS 2024 and IEEE VCIP 2018.  He received two best paper awards from VCIP 2010 and 2015. He is serving as an Associate Editor-in-Chief and a Senior Area Editor of \textsc{IEEE Transactions on Circuits and Systems for Video Technology}. He was an Associate Editor of \textsc{IEEE Transactions on Image Processing}, \textsc{IEEE Transactions on Circuits and Systems for Video Technology}, \textsc{IEEE Transactions on Multimedia}, and \textsc{IEEE Multimedia}. He received a Distinguished Research Award presented by National Science and Technology Council, Taiwan, in 2023.
\end{IEEEbiography}

\vspace{-0.4in}
\begin{IEEEbiography}
	[{\includegraphics[width=1in,height=1.25in,clip,keepaspectratio] {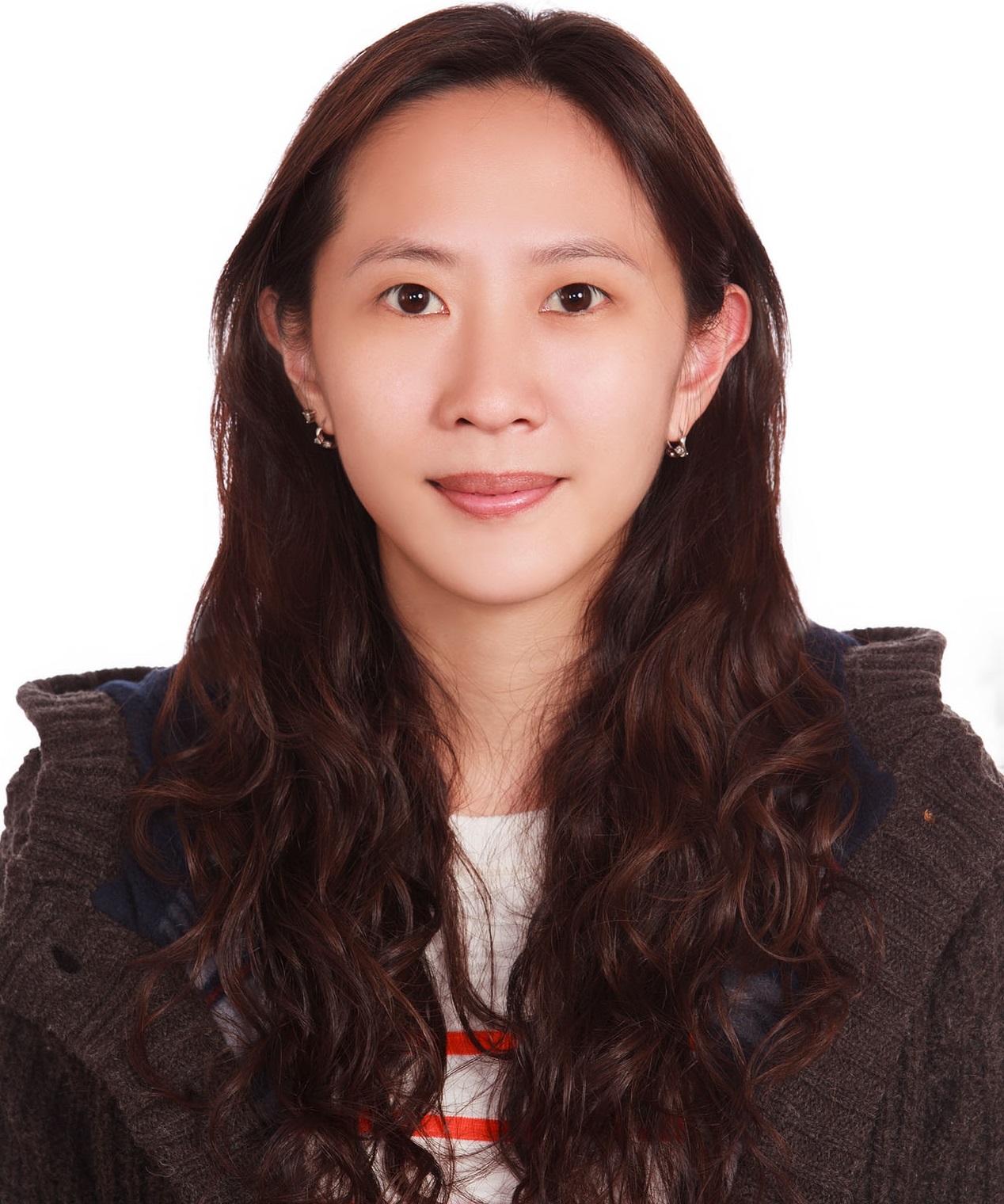}}]
	{Shao-Yun Fang}
	(Member, IEEE) received a B.S. degree in electrical engineering from the National Taiwan University (NTU), Taipei, Taiwan, in 2008 and a Ph.D. degree from the Graduate Institute of Electronics Engineering, NTU, in 2013. She is currently a Professor of the Department of Electrical Engineering, National Taiwan University of Science and Technology, Taipei, Taiwan. Her research interests focus on physical design and design for manufacturability for integrated circuits. Dr. Fang received two Best Paper Awards from the 2016 International Conference on Computer Design and the 2016 International Symposium on VLSI Design, Automation, and Test, and two Best Paper Nominations from  International Symposium on Physical Design in 2012 and 2013.
\end{IEEEbiography}

\vspace{-0.4in}

%\vspace{-0.4in}
\begin{IEEEbiography}
	[{\includegraphics[width=1in,height=1.25in,clip,keepaspectratio] {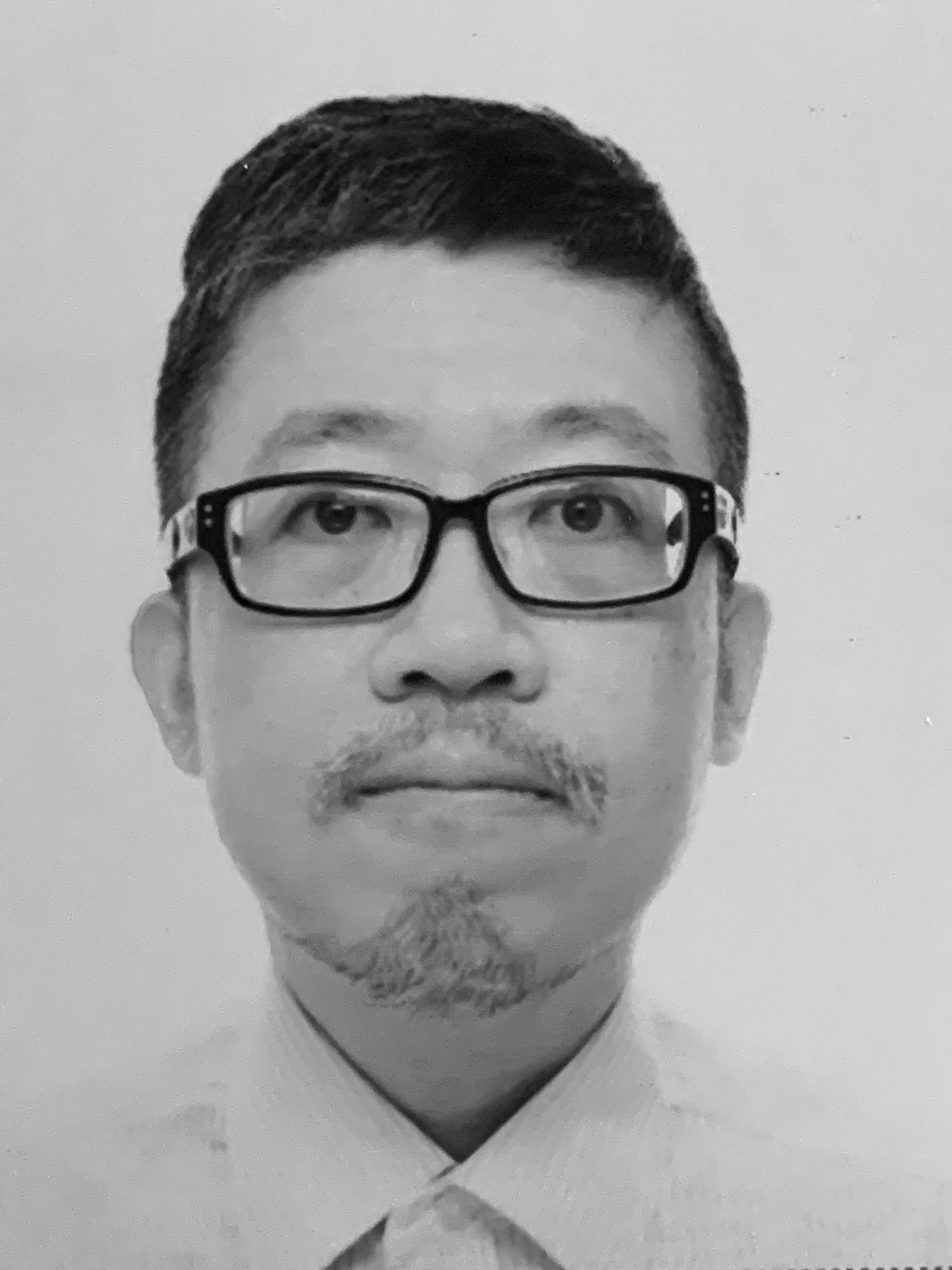}}]
	{Yan-Hsiu Liu}
	received his M.S. degree in Chemistry from National Tsing Hua University (NTHU), Taiwan, in 2002. In 2004, he joined United Microelectronics Corporation (UMC) as a process integration engineer in Hsinchu, Taiwan. He is currently working as a deputy department manager on the development of smart manufacturing and responsible for industry-academia cooperation/collaboration. His research interests include the areas of intelligent manufacturing systems, adaptive parameter estimation, and neural networks.
\end{IEEEbiography}